%% file: main.tex
\pdfoutput=1
\documentclass[11pt]{article}

\usepackage[]{acl}

\usepackage{times}
\usepackage{latexsym}

\usepackage[T1]{fontenc}
\usepackage[utf8]{inputenc}

\usepackage{microtype}
\usepackage{graphicx}
\usepackage{amsmath}
\usepackage{multirow}
\usepackage{booktabs}
\usepackage{subcaption}
\usepackage{enumitem}
\usepackage{array}
\newcolumntype{?}{!{\vrule width 1pt}}
\usepackage{comment}
\usepackage{makecell}
\usepackage{subfiles}
\usepackage{amsfonts}

\usepackage{mathtools,tikz}
\DeclareRobustCommand\sampleline[1]{%
  \tikz\draw[#1] (0,0) (0,\the\dimexpr\fontdimen22\textfont2\relax)
  -- (1.5em,\the\dimexpr\fontdimen22\textfont2\relax);%
}

\definecolor{dashcolorred}{HTML}{A62A17}

\title{CaM-Gen: Causally Aware Metric-Guided Text Generation}

\author{Navita Goyal\\
  University of Maryland\thanks{ Work done while at Adobe Research}\\
  \texttt{navita@umd.edu} \\\And
  Roodram Paneri \\
  Microsoft$^*$ \\
  \texttt{rpaneri@microsoft.com} \\\And
  Ayush Agarwal \\
  Tournafest$^*$ \\
  \texttt{aagarwal9782}\\
  \texttt{@gmail.com}\AND
  Udit Kalani \\
  Adobe Systems$^*$\\
  \texttt{kalani@adobe.com} \\\And
  Abhilasha Sancheti \\
  University of Maryland\\
  \texttt{sancheti@umd.edu} \\\And
  Niyati Chhaya \\
  Adobe Research \\
  \texttt{nchhaya@adobe.com} \\
  }

\begin{document}
\maketitle
\begin{abstract}
Content is created for a well-defined purpose, often described by a metric or signal represented in the form of structured information. The relationship between the goal (metrics) of target content and the content itself is non-trivial. While large-scale language models show promising text generation capabilities, guiding the generated text with external metrics is challenging.
These metrics and content tend to have inherent relationships and not all of them may be of consequence. We introduce CaM-Gen: Causally aware Generative Networks guided by user-defined target metrics incorporating the causal relationships between the metric and content features. We leverage causal inference techniques to identify causally significant aspects of a text that lead to the target metric and then explicitly guide generative models towards these by a feedback mechanism. We propose this mechanism for variational autoencoder and Transformer-based generative models. The proposed models beat baselines in terms of the target metric control while maintaining fluency and language quality of the generated text. To the best of our knowledge, this is one of the early attempts at controlled generation incorporating a metric guide using causal inference.
\end{abstract}

\subfile{sections/introduction}
\subfile{sections/related-work}
\subfile{sections/methods}
\subfile{sections/experiments}
\subfile{sections/conclusion}
\subfile{sections/ethics}

\bibliography{custom}
\bibliographystyle{acl_natbib}

\appendix

\subfile{sections/appendix}

\end{document}

%% file: sections/introduction.tex
\section{Introduction}
Most content is created for a well-defined goal. For example, a blog writer often publishes articles to gain popularity and trigger conversations, and a columnist may write an opinionated piece to gather feedback. In marketing applications, these goals are business objectives that need to be optimized using the content shared with the customers. The validation of whether the goal was met or not is done by tracking metrics that capture the reader behavior. In social media, metrics include number of comments, likes, or shares whereas for a publishing house they are the number of views and readers. These engagement metrics (hereafter, metrics) are proxy for target goals. Based on historical content, textual content characteristics that successfully achieve the desired metrics can be assessed \cite{tan-2019-user-response, verma-2020-modelling}. Guiding text generation models by these signals is important for meeting the required goals.
% The challenge in such goal-specific content, where the goal is defined by a metric independent of the content features, is to first identify the right relationship between metric and content and then to have an appropriate generative model that is informed by this metric as a target guide. 
% We introduce methods for metric-guided text generation achieved by (a) metric and content-specific losses in a conditional variational autoencoder and (b) by modifying the attention and layer normalization layers in a Transformer-based generator.

While recent neural language models have shown tremendous success towards fluent text generation \cite{radford-wu-2018-gpt-2, devlin-etal-2019-bert}, achieving controlled, goal-specific generation is challenging. There has been work on text generation controlling for style, topic, or size \cite{keskar2019ctrl}. These methods are able to leverage content characteristics that are common between the definition of goal (i.e., control) and the text. However, for metrics that are not explicit in the text, controlled generation is non-trivial to codify. 
% The challenge is introduced due to the fact that the metric is independent of the content representation space, hence creating a need to first identify the relationship between the content characteristics and the metric and then to explicitly introduce a guide/constraint enabling the generator to learn the desired content properties. Contrary to style, these choices might be difficult for a layman to manually identify and input to the generative models.
The challenge is introduced due to the fact that for external metrics, there is a need to first identify the relationship between the content characteristics and the metric and then to explicitly introduce a guide/constraint enabling the generator to learn the desired content properties. Contrary to style, these choices might be difficult for a layman to manually identify and input to the generative models.

% We introduce methods for metric-guided text generation achieved by (a) metric and content-specific losses in a conditional variational autoencoder and (b) by modifying the attention and layer normalization layers in a Transformer-based generator.
%As mentioned earlier, the metric and the content are disjoint in their representation spaces. There is hence a need to codify the relationship between various content features with the metric (i.e the outcome variable) to ensure that the generative model is directed towards the appropriate features for the generation. 
Textual content is an amalgam of various linguistic features --- lexical, pertaining to word choices; semantics, concerned with the meaning; syntactic, relating to parts of speech tags; and surface-level features, comprising punctuation, word count, sentence count, etc. 
% Textual content is an amalgam of various semantic, syntactic, surface, and lexical features \cite{2019-verma-linguistic}.
% to more external meta-features that qualify the type or author. 
% As is expounded in causal literature, a correlation analysis between these features and the target outcome is insufficient \cite{aldrich1995}. 
To avoid misinformation (or clickbait-y) generation, automated tools should be able to alter the syntactic and surface-level characteristics of text to meet the desired outcome. Explicitly identifying features of interest that result in intended outcome can enable finer control. 
In this paper, we first discuss method to identify a subset of these features that have direct and significant impact on the outcome metric, derived from causality literature \cite{10.1093/aje/kwq439}. 
% We then develop generation models guided with these causal features enabling finer control.
% While a large range of these features may have a non-zero correlation with the metric, it is highly unlikely that the features and the data hold a causal relationship. 
A causally significant relationship helps encode the `if this, then that' logic; adding such a guide for the generator can help ensure on-metric generation. 

In this paper, we propose causal guidance mechanism for two modeling frameworks that are used for metric-guided generation --- conditional variational autoencoders \cite{sohn-2015-learning} and Transformer-based language models \cite{vaswani-2017-Transformers}. 
% We introduce metric guidance by adding specific losses and by modifying self-attention and layer normalization in Transformer blocks.
% We introduce `causal losses' in our models to incorporate a guidance along causally significant features, identified through an analysis that calculates the average treatment effect between the metric and the content features. 
For conditional variational autoencoders (CVAE), we modify the VAE graph to introduce causal guidance. In Transformer-based language models, we introduce causal guidance by adding causal losses for explicit feedback on causal features. 

Our key contributions are introducing causal guidance frameworks for metric-guided, controlled text generation in CVAE and Transformer-based generative models. We experiment with a new dataset of news articles related to COVID-19 along with the NYT-comments dataset,\footnote{\url{https://www.kaggle.com/aashita/nyt-comments}} showing improved performance against baseline methods. To the best of our knowledge, this is one of the first attempts towards controlled generation on engagement metrics and inclusion of causal guidance for controlled generation in generative models. 

%% file: sections/related-work.tex
\section{Related Work}
The literature on text generation spans various generative models, including variational autoencoder (VAEs), generative adversarial networks (GANs), and sequential models. VAEs have been used for unconditional \cite{bowman-etal-2016-generating}, as well as constrained text generation \cite{zhang-etal-2016-variational-neural, pagnoni2018conditional}. 
% While historically used in computer vision, VAEs are increasingly adapted to text domain \cite{zhang-etal-2016-variational-neural, wang-etal-2019-topic-guided}, specially for applications that can leverage the probabilistic model structure. To enable constrained generation, conditional VAE (CVAE) are implemented in image \cite{ yan2015attribute2image} and text domain \cite{sohn-2015-learning, pagnoni2018conditional}.
\citet{pagnoni2018conditional} generate a sentence sequence $y$ conditioned on the input sentence for machine translation, thus mimicking a sequence-to-sequence model. 
% These methods do not allow control over an external attribute or metric.
\citet{zhiting-2017-towards} control sentiment and tense in text generation using discriminators with VAEs. \citet{zhao-etal-2017-kgcvae} introduce an additional reconstruction network in CVAEs for controlling linguistic features in dialog generation.
% \citet{zhao-etal-2017-kgcvae} control linguistic features in dialog generation by incorporating an additional reconstruction network in CVAEs. 
% This enables mimicking the linguistic characteristics in dialog. 
As we show in our experiments, this does not adapt well to controlled generation where the relationship with the target goal is not as explicit in text. 
We identify these nuanced relationships between the text and the underlying goal and enable explicit control over the text features influencing the target outcome by modifying the VAE graph.
% However this does not allow for closer control over specific aspects of the text that affect the target metric. We incorporate this by introducing causal graph to the graphical model of VAEs, which enables the feedback and control for finer features of text that lead to the target outcome.

While VAEs enable controlled generation, they do not generate fluent language with limited data. Large Transformer-based language models \cite{radford-wu-2018-gpt-2, devlin-etal-2019-bert} have shown efficacy in generating fluent language, allowing for fine-tuning for specific tasks on a smaller dataset while maintaining good language quality. 
% There have been significant efforts in introducing control over generation in these language models. 
\citet{keskar2019ctrl} introduce style control, such as domain (books, wikipedia, etc.), by conditioning the generated distribution on the style token $y$, i.e. $p(x|y)=\prod_{i=1}^n p(x_i|x_{<i}, y)$. The language model learns the conditional probability $p(x_i|x_{<i}, y)$ by training on sequences of raw text prepended with the style control. This approach provides only weak control, especially if the variation in textual features for the same target metric is large. \citet{zeng2020style} enable finer control over generation space by introducing the control $y$ in various internal layers of the Transformer network. \citet{singh-etal-2020-incorporating} control for a combination of lexical styles to reproduce author's styles using a RL framework for Transformer-based language models.
While style is well reflected in the choice of vocabulary and language distribution, the difference in the language distribution is not as apparent for an external metric as control. We observe that the external metric is more influenced by various syntactic and surface-level text features, as opposed to the underlying vocabulary. We achieve finer control over these by a causally aware generative language model.
% Ideally such text generation should be able to perform controlled generation for external control metric (i.e. goal). However, as opposed to style, the syntactic and lexical features are more diverse in text with similar target metric. Although the engagement is dictated by the text, it manifests very subtly in  textual features.
% Although the same can be adapted for an external control metric, this approach lacks finer control. The syntactic and lexical features, as against style, in text are more diverse for across text inspite similar target metric. 
% In such cases, a weak control (such as, prepending text stream with metric) is unlikely to be sufficient in learning nuances of the relationship between the metric and the content. In contrast, we add this control in various internal layers via self-attention and layer normalisation introducing a finer metric control in the generative model.

% \citet{zeng2020style} introduce a similar notion for style control in language models, allowing finer control for target styles. While style is well reflected in the choice of vocabulary or the language distribution, the difference in the language distribution is not as apparent for an external metric as control. As observed through our analysis, the external metric is more influenced by various lexical and syntactic text features, as opposed to the underlying vocabulary. To this extent, the style control models are limited in their capabilities beyond the choice of language. We overcome this by introducing a causally-aware generative language model.

\textbf{Causal Inference.} Causal analysis entails dissecting the effects of specific treatment on the outcome variables, while controlling for other confounding factors. These methods are widely used in fields such as marketing, advertising, healthcare and more recently textual analysis \cite{feder2021causal}.
% Recent works explore the applicability of these methods in textual space. 
Causal inference in text has many facets, as expounded in \citet{feder2021causal}. In this work, our focus is understanding the effect of specific characteristics of text on the outcome of interest. 
Previous work in this area has studied various text characteristics and outcomes, such as effect of words on sentiment classification \cite{paul-2017-feature}, effect of presence of theorems on the acceptance rate of papers and the effect of gender on the popularity of social media posts \cite{veitch2019adapting}, and the effect of specific content features on the user response \cite{tan-2019-user-response, verma-2020-modelling}. 
% \citet{paul-2017-feature} employ a propensity matching algorithm to identify causal association between the text with its sentiment classification. \citet{veitch2019adapting} study the effect of presence of elements such as theorems on the acceptance rate of papers or the effect of gender on the popularity of social media posts. 
These work focus on identifying the effect of textual features on the outcome. We go one step further and aim at introducing causal guidance in text generation. 
% \citet{verma-2020-modelling} use a doubly robust method on propensity-based matching to estimate the causal effects. They use adaptive and flexible multi-layer neural networks to model potential outcomes. We adapt their technique to uncover causal effect of various syntactic and surface-level features in textual data and then use these for guiding causally-aware generation. 
% , as described in detail in the following sections.

%% file: sections/methods.tex
%\section{Approach}
%Our approach has 2 key components: the identification of causally significant features in text and guiding the generative models towards generating relevant causal aspects. For the generative mode, both a VAE-based model and a Transformer-based architecture are considered.

%We consider a CVAE architecture to condition the generation with the target metric and then introduce the causal graph to incorporate causally significant features into the generated text. For the Transformer generative model, we condition on the target outcome metric by modifying several layers of the Transformer network and incorporate the causal aspect in form of feedback over the generated content.

\section{Causal Features Identification}\label{section:causal-effect}
To incorporate finer control over generation of text to achieve a specific target metric, we first identify features that contribute to the respective outcome. Here, the outcome metric is the target value we wish to control. We consider various syntactic (e.g. \textit{noun/adjective count}) and surface-level textual features (e.g. \textit{word/sentence/paragraph count}) and measure their effect on the metric. Consider two text choices -- S1: ``\textit{The dog sprinted ahead so fast, the girl had much hard time keeping up with it.}", S2: ``\textit{The dog sprinted fast ahead. The girl panted trying to keep up.}''; both meaningful and reasonable generations. Say, textual content with less words per sentence and more sentences is better liked. In this case, \textit{word count} would have negative effect on outcome metric and \textit{sentence count} would have a positive effect. Thus, the model should generate shorter sentences, resulting in S2. Although this example uses semantically equivalent text pieces for illustration, we do not have such parallel instances for generation task discussed in the paper. In absence of parallel data, it is non-trivial to isolate the effect of a specific text feature on the outcome metric. 
% The difference here is not as much in lexical or semantic, but the surface-level elements that might appeal to audience
Thus, we turn to causal estimation methods to identify this effect without controlled parallel data. 

The hypothetical change in an input feature of text in the observed data is defined as an intervention, and the input feature in question is termed as the \textit{treatment variable} ($t$). For a binary treatment, the effect of treatment on the outcome (i.e., $y$) in the $i^{t h}$ text sample is defined as $y_1(x_i)-y_0(x_i)$. Here, $y_0$ represents outcome in absence of treatment and $y_1$ represents outcome when treatment is applied and $x_i$ are the other covariates (text features). The average treatment effect (ATE) is the expected effect of providing the treatment (i.e. including a specific feature) and is given by $\mathbb{E}[y_1(x_i)-y_0(x_i)]$. 
This can not be directly calculated as 
% interventions are not observed in data, i.e. 
we do not know what the outcome is if a certain part of text is changed in a certain way, i.e., $y_0(x_i)$ and $y_1(x_i)$ is not known for the same $i$.
% The challenging aspect of this analysis is that interventions are not observed in data, i.e. we do not know what the outcome is if a certain part of text is tweaked in a certain way. Thus, we do not observe $y_0(x_i)$ and $y_1(x_i)$ for the same $i$.
Moreover, in observed data, the treatment assignment is not independent of baseline covariates. We account for this by employing a propensity-based scoring, which serves to balance treatment assignment in treated and untreated groups \cite{doi:10.1080/00273171.2011.568786}.
% As a result, the intrinsic characteristics of the treated group are inherently different from the untreated group. 
% We account for these systematic differences while estimating effect of treatment by employing a propensity-based scoring.
% We need to account for these systematic differences while estimating effect of treatment on outcome metric. We employ a propensity-based scoring method. 

The propensity score is defined as the probability of treatment assignment conditional on baseline covariates, i.e. $\pi(x_i)$ = $p(t_i=1|x_i)$. 
% The propensity score acts as a balancing score, that is, conditional on propensity score the distribution of baseline covariates is similar in treated and untreated group \cite{doi:10.1080/00273171.2011.568786}. 
% As is expounded in causal literature, a correlation analysis between \textit{treatment} and \textit{outcome} variables is not sufficient, since there are many other features in the text that would have an effect on the outcome, and a correlation analysis will be insufficient in separating the effect of treatment variable from these other features. These features are formally known as \textit{confounders}. To account for the confounding variables, we employ a propensity-based scoring method to identify this causal effect \cite{tan-2019-user-response}. 
We employ multi-layer neural networks to approximate propensity scores \cite{tan-2019-user-response}. The propensity scoring model is trained using the assigned treatment $t_i$ corresponding to the observed covariates $x_i$ with cross entropy loss. 
% We train a binary class neural network $F_b(.)$, which estimates propensity score of $i^{th}$ text sample as $\pi(x_i)$ = $F_b(x_i; \theta_b)$. Here, $\theta_b$ are model parameters trained on minimizing binary cross entropy loss for prediction of $t_i$, i.e. $\mathcal{L}$ = $-\frac{1}{n}\sum_{i=1}^n t_i\log\pi(x_i)$ + $(1-t_i)\log(1-\pi(x_i))$. 
The average treatment effect (ATE) can be estimated by inverse propensity treatment weighing (IPTW) \cite{doi:10.1080/00273171.2011.568786}, where each outcome is weighed by inverse probability of receiving the corresponding treatment. Thus,
\begin{equation}
% \small
    ATE=\frac{1}{n}\sum_{i=1}^n \bigg[\frac{t_i y_i}{\pi(x_i)} - \frac{(1-t_i) y_i}{1-\pi(x_i)}\bigg]
\end{equation}
% This estimator is unbiased if the propensity score model is correctly specified. To account for inadequacies in propensity scoring model, we use a doubly robust estimation where the IPTW is augmented with potential outcome estimator \cite{10.1093/aje/kwq439}. The potential outcome models estimates the outcome if treatment is applied $(t$=$1)$ or not applied $(t$=$0)$, given the other covariates. We model potential outcome using two neural networks (for $t$=$0,1$), trained to minimize mean squared error in predicted and actual outcome in observed articles with $t$=$1$ and $t$=$0$, respectively. 
For a doubly robust estimate, we augment IPTW with potential outcome model \cite{10.1093/aje/kwq439}. The potential outcome models estimate outcomes if treatment is applied $(t=1)$ or not applied $(t=0)$, given the other covariates. We model potential outcome using two neural networks (for $t=0,1$), trained to minimize mean squared error in predicted and actual outcome in observed articles with $t=1$ and $t=0$, respectively. 
% for $t$=$\{0,1\}$, given by, $F_o^t (x_i; \theta_o^t )$ = $\mathbf{E}[y|t_i$ = $t,x_i ]$ = $\hat{y}_t(x_i)$. $F_o^{t=1}$ and $F_o^{t=0}$ are trained to minimize mean squared error in predicted and actual outcome in observed articles with $t$=$1$ and $t$=$0$, respectively. 
The expected outcome in presence of the treatment feature is then a function of the observed outcome with treatment for the treated group and the predicted outcome with treatment for the untreated group, given article features, weighted by a function of the propensity scores. 
% More specifically, 
\begin{equation}
% \small
    y_1(x_i)=\frac{t_i y_i}{\pi(x_i)}-
    \frac{t_i-\pi(x_i)}{\pi(x_i)}\hat{y}_1(x_i)
\end{equation}
Similarly, the overall response in the absence of treatment is estimated as
\begin{equation}
    y_0(x_i)=\frac{(1-t_i) y_i}{1-\pi(x_i)}+
    \frac{t_i-\pi(x_i)}{1-\pi(x_i)}\hat{y}_0(x_i)
\end{equation}
The average effect of the treatment feature on the outcome is estimated as the mean of the difference of expected outcome with and without treatment.
\begin{equation}
    ATE=\frac{1}{n}\sum_{i=1}^n (y_1(x_i)-y_0(x_i))
\end{equation}
% As shown in appendix, these formulations give doubly robust unbiased estimate of average treatment effect. 
This provides an estimate of which text features have the most impact on the outcome (target) metric.\footnote{Table \ref{tab:causal-accuracy} lists the features discussed in the paper. Complete list of features and their ATE is included in Appendix D} The ATE of continuous treatment features can be estimated in a similar fashion, assuming a normal treatment distribution \cite{tan-2019-user-response}.
% The ATE helps estimate which text features have the most impact on the outcome (target) metric for a given piece of text. 
% We calculate ATE for various features, like \textit{word count}, \textit{sentence count}, \textit{paragraph count} and parts of speech tags, like \textit{noun count}, and \textit{adjective count}.\footnote{List of features and their ATE is available in Appendix D.} 

% These insights are incorporated into the model to lead the generated text to adopt these features and achieve the target outcome.

\section{CaM-Gen}
We present a causally aware text generation method in VAE and Transformer-based models. In section \ref{section:CVAE}, we begin by discussing the metric-guided generation framework in Variational Autoencoders (VAE) \cite{zhang-etal-2016-variational-neural}. We then describe our causal-guidance mechanism which augments this conditional VAE (CVAE) with a causal graph to incorporate causally significant features in generative process. In Transformer-based text generation (section \ref{section:Transformer}), we first discuss controlled text generation by modifying Transformer layers with respect to the target control \cite{zeng2020style}. We then introduce our proposed causal feedback mechanism to guide the model towards pre-identified causal features for controlled generation. We conclude with section \ref{parallel} comparing and drawing parallels between the two generative frameworks and their respective causal mechanisms.
% We present causally-aware text generation method in VAE and Transformer-based models. We consider a CVAE architecture for conditional generation with target metric (section \ref{section:CVAE}) and augment VAE with causal graph to incorporate causally significant features in text generation (section \ref{section:causalCVAE}). In Transformer-based architecture, we condition on target metric by modifying several layers of the Transformer network (section \ref{section:Transformer}) and then introduce causal feedback process for causal guidance in controlled text generation (section \ref{section:causalTransformer}).

\begin{figure}[t]
    \centering
    \includegraphics[width=0.45\textwidth]{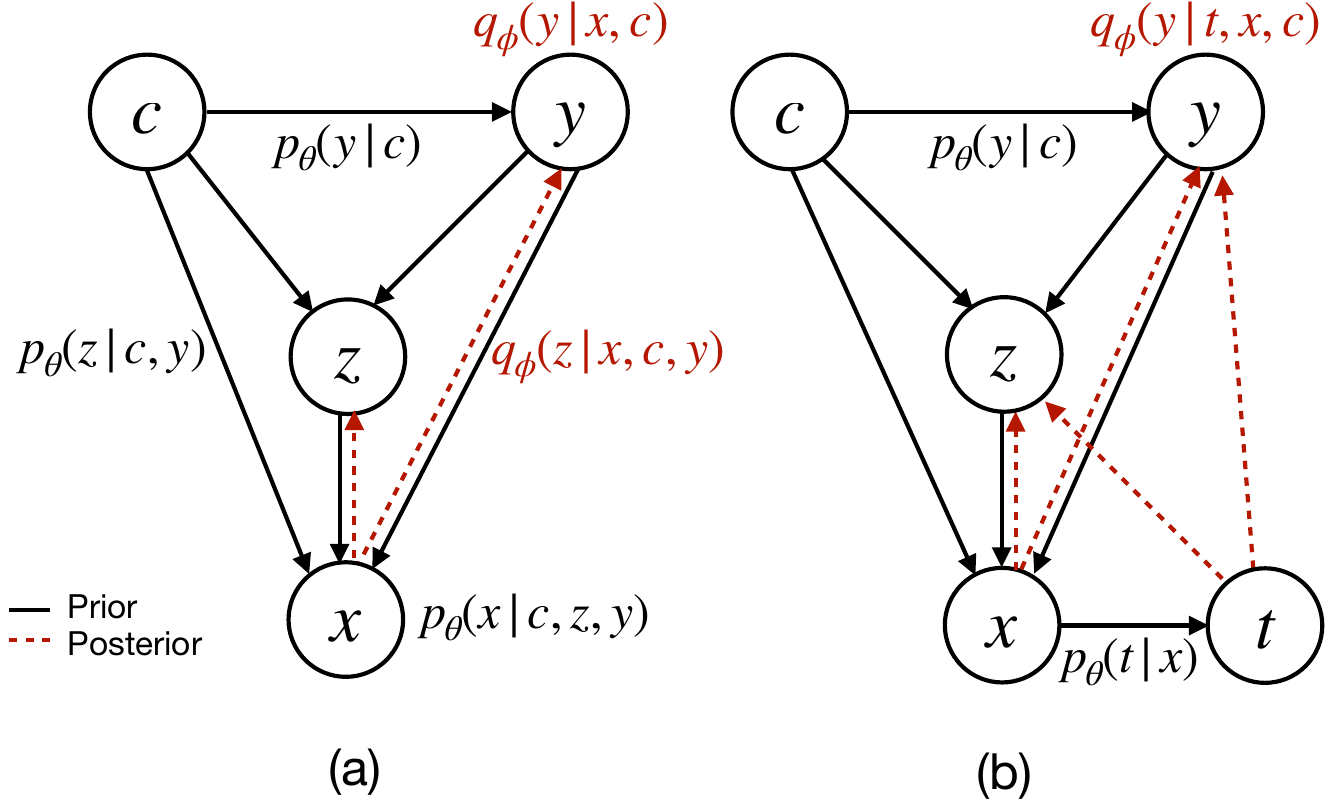}
    \caption{VAE Graph - (a) Conditional generation, (b) Causal feedback in conditional generation. Black solid line (\textbf{\sampleline{}}) and red dashed line (\textbf{\textcolor{dashcolorred}{\sampleline{densely dashed}}}) corresponds to the prior and posterior network connections}
    \label{fig:causal-graph}
\end{figure}

% \begin{figure}
%     \centering
%     \begin{subfigure}[t]{0.18\textwidth}
%     \centering
%         \includegraphics[scale=0.25]{Images/causal_graph_1.pdf}
%         \caption{CVAE}
%         \label{fig:baseline}
%     \end{subfigure}
%     \begin{subfigure}[t]{0.18\textwidth}
%     \centering
%         \includegraphics[scale=0.25]{Images/causal_graph_2.pdf}
%         \caption{Causal CVAE}
%         \label{fig:neither}
%     \end{subfigure}
%     \caption{VAE Graph}
%     \label{fig:causal-graph}
% \end{figure}

\subsection{Conditional Variational Autoencoder\label{section:CVAE}}
We first adapt the CVAE architecture, inspired by \citet{zhao-etal-2017-kgcvae}. As opposed to generating a response to previous utterances, we model the conditional generation as a next sentence generation task -- generate the next sentence $x$, given the previous context $c$, and the target metric $y$.
% Instead, we estimate the prior of $y$ as $p_\theta(y|c)$.
% Instead, we consider outcome metric $\tilde{y}$ as a substitute for the actual target metric $y$. 

% Consider a latent variable $z$, that captures the latent distribution over the generation space. To condition the generation on the target metric, we try to reconstruct $y$ from the distribution of $p(\tilde{y}=y|c,z)$. The generated sentence $x$ is thus dependent on $c$, $z$ and $\tilde{y}$. This leads to an explicit control over the target metric, while also accounting for the role of context in affecting the outcome metric. The graph for this is as shown in Fig.\ref{figure-causal-graph}a.
% We construct the prior network $p(z|c)$ to estimate the latent variable $z$, assuming a multi-variate Gaussian distribution. To estimate the outcome metric $\tilde{y}$, we have a reconstruction network for $p(\tilde{y}|c,z)$. Finally, a decoder network is used for generation, given by $p_\theta(x|c,z,\tilde{y})$. We consider a recognition network $q_{\phi}(z|x,c,\tilde{y})$ to approximate the true posterior $p_\theta(z|x,c,\tilde{y})$. The CVAE network can be trained using the variational lower bound given by
We consider a latent variable $z$ that captures the latent distribution over the generation space. We estimate $z$ using the prior network $p(z|c, y)$, assuming a multi-variate Gaussian distribution. The sentence $x$ is generated by the decoder network $p_\theta(x|c,z,y)$.
% To condition the generation on the target metric, we try to reconstruct $y$ from the distribution of $p(\tilde{y}=y|c,z)$. 
% The generated sentence $x$ is thus dependent on $c$, $z$ and $\tilde{y}$. A decoder network is used for generation, given by $p_\theta(x|c,z,\tilde{y})$. 
% This leads to an explicit control over the target metric, while also accounting for the role of context in affecting the outcome metric. 
% To estimate the outcome metric $\tilde{y}$, we have a reconstruction network for $p(\tilde{y}|c,z)$. 
The prior of the outcome metric is approximated using $p_\theta(y|c)$. Since the outcome metric depends on both the generated $x$ and the given context $c$, we do not assume independence between the inputs $c$ and $y$. We consider two recognition networks $q_{\phi}(y|x,c)$ and $q_{\phi}(z|x,c,y)$
to approximate the true posteriors $p_\theta(y|x,c)$ and $p_\theta(z|x,c,y)$ (graph as shown in Fig. \ref{fig:causal-graph}a). The CVAE network can be trained using the variational lower bound.\footnote{Proof included in Appendix A.1}

\begin{equation}
\begin{split}\label{eq:cvae-non-causal}
    \mathcal{L}_{\operatorname{V_{nc}}}(\theta, &\phi; x,c,y) =\mathbb{E}_{q_\phi(z,y|x,c)}[\log p_\theta(x|c,z,y)] \\
    &-\mathbb{E}_{q_\phi(y|x,c)}\operatorname{KL}[q_{\phi}(z|x,c,y)||p_\theta(z|c, y)]\\
    &-\operatorname{KL}[q_{\phi}(y|x,c)||p_\theta(y|c)]
\end{split}
\end{equation}
Intuitively, the first term is the reconstruction loss; the second term aligns the latent variable $z$ with respect to the metric $y$ and the generated text $x$; and the last term ensures that generation adheres to the target metric. 
% \begin{equation}
% \small
% \begin{split}\label{eq:cvae-non-causal}
%     \mathcal{L}_{\operatorname{V_{nc}}}(\theta, \phi; x,c,y) &=\mathbf{E}_{q_\phi(z|x,c,y)}[\log p_\theta(x|c,z,y)] \\
%     &-\operatorname{KL}[q_{\phi}(z|x,c,y)||p_\theta(z|c, y)] \\
%     &-\operatorname{KL}[q_{\phi}(y|x,c)||p_\theta(y|c)]
% \end{split}
% \end{equation}
% \begin{equation}
% \small
% \begin{split}\label{eq:cvae-non-causal}
%     \mathcal{L}_{\operatorname{V_{nc}}}(\theta, \phi; x,c,\tilde{y}) &=-\operatorname{KL}(q_{\phi}(z|x,c,\tilde{y})||p_\theta(z|c)) \\
%     &+ \mathbf{E}_{q_\phi(z|x,c,\tilde{y})}[\log p_\theta(x|c,z,\tilde{y})] \\
%     &+ \mathbf{E}_{q_\phi(z|x,c,\tilde{y})}[\log p_\theta(\tilde{y}=y|c, z)]
% \end{split}
% \end{equation}

\textbf{Causal-guidance in CVAE. } \label{section:causalCVAE}
The above conditional generation controls the target metric as a whole, but does not directly influence specific aspects of the text that impact the outcome metric. Ideally, the latent variable $z$ would implicitly learn these during training. However, in practice this is not so, especially in the case of limited data and multiple confounders. 
% In its entirety, the generated text $x$ comprises of several contextual and semantic information. 
Besides aligning the latent space $z$ w.r.t. $x$, we enable explicit causal guidance by aligning the latent space to the \textit{causally significant features} $t$ (features significantly impacting the target metric) in the generated text. Causal feature vector $t$ comprises features with ATE (section \ref{section:causal-effect}) higher than a threshold.\footnote{Significance threshold are chosen empirically. See Causal Feature Identification in Section \ref{section:results} for details}

% Consider the causally significant features in text, identified using the ATE analysis (section \ref{section:causal-effect}), as the treatment vector $t$. For instance, if \textit{word count} has significant ATE, then the vector $t$ would constitute the word count of the generated text $x$. 
The posterior distribution of latent variable $z$ is now estimated as $q_\phi(z|t,x,c,y)$. By definition, the outcome metric distribution will be affected by the causal features $t$ in the generated $x$. The posterior distribution for outcome metric $y$  can hence be approximated as $q_\phi(y|t,x,c)$. The feedback of these causal effects is propagated through the network by minimizing the KL divergence between the prior distribution $p_\theta(y|c)$ and $q_\phi(y|t,x,c)$ (Fig. \ref{fig:causal-graph}b). The loss function\footnote{Proof included in Appendix A.2} for causal CVAE is
% \begin{equation}\label{eq:cvae-causal}
% \small
% \begin{split}
%     \mathcal{L}(\theta, \phi; x,c,\tilde{y}) = &-KL(q_{\phi}(\tilde{y}|t,x,c)||p_\theta(\tilde{y}|c,z)) \\
%     &+ \mathbf{E}_{q_{\phi}(\tilde{y}|t,x,c)}[\log p_\theta(t|x)] \\
%     &- KL(q_{\phi}(z|x,c,\tilde{y})||p_\theta(z|c)) \\
%     &+ \mathbf{E}_{q_\phi(z|x,c,\tilde{y})}[\log p_\theta(x|c,z,\tilde{y})] \\
%     &+ \mathbf{E}_{q_\phi(z|x,c,\tilde{y})}[\log p_\theta(\tilde{y}=y|c, z)]
% \end{split}
% \end{equation}
\begin{equation}\label{eq:cvae-causal}
% \small
\begin{split}
    \mathcal{L}_{\operatorname{V_{c}}}(&\theta, \phi; t,x,c,y) 
    =\\
    &\mathbb{E}_{q_\phi(z,y|t,x,c)}[\log p_\theta(x|c,z,y)] \\
    -&\mathbb{E}_{q_\phi(y|t,x,c)}\operatorname{KL}[q_{\phi}(z|t,x,c,y)||p_\theta(z|c, y)] \\
    +&\mathbb{E}_{q_{\phi}(z,y|t,x,c)}[\log p_\theta(t|x,c,z,y)]\\
    -&\operatorname{KL}[q_{\phi}(y|t,x,c)||p_\theta(y|c)]
\end{split}
\end{equation}
% where $\mathcal{L}_{V_{nc}}$ is loss for non-causal CVAE (Eq. \ref{eq:cvae-non-causal}).

\begin{figure}[t]
    \centering
    \includegraphics[width=0.45\textwidth]{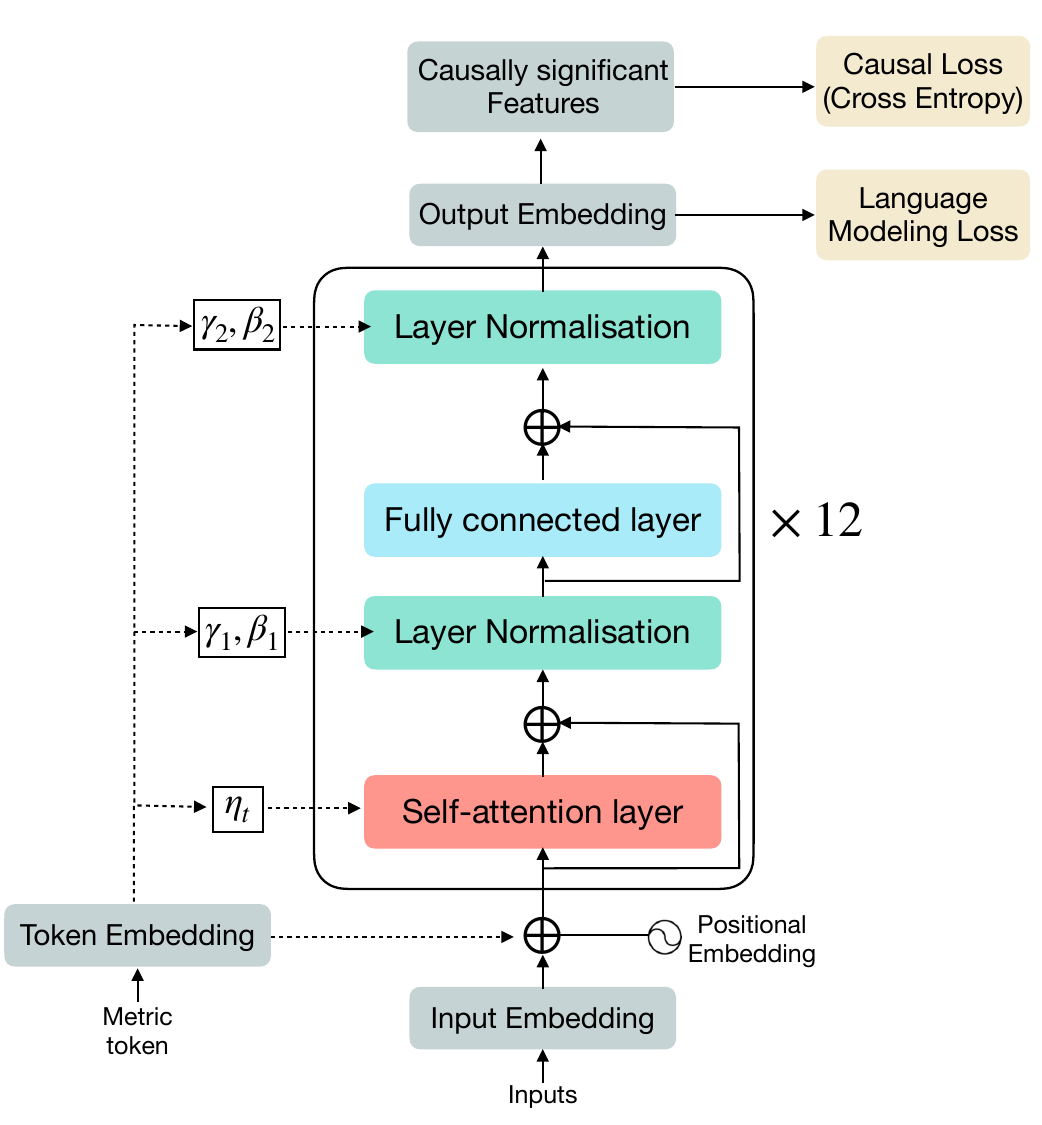}
    \caption{CaM-Gen: Transformer}
    \label{figure-Transformer-architecture}
\end{figure}

\subsection{Conditional generation in Transformer \label{section:Transformer}}
The proposed Transformer model is based on the GPT-2 architecture \cite{radford-wu-2018-gpt-2}, which is trained on language modeling loss for predicting the next token given all the previous tokens. The model is first pre-trained with language modeling objective on a large corpora to build understanding of language distribution enabling it to generate coherent text. Although fine-tuning with the same objective shifts the language distribution of generated text towards the fine-tuning corpus, explicitly controlling for a target metric is more nuanced. 
To introduce this explicit control, we use the metric to modify self-attention and normalization layers in the Transformer blocks \cite{zeng2020style}, as shown in Fig.~\ref{figure-Transformer-architecture}.\footnote{$\eta$, $\gamma$, $\beta$ are the scale/bias parameters in respective layers (details in Appendix B)} In the former, attention weights of Transformer blocks are biased towards the target by changing the query vector in attention mechanism with the affine transformation of $y$. In the latter, the scale and bias parameters of layer normalization are replaced by functions of $y$. This ensures that the target information does not \textit{wash away} \cite{park2019normalisation} and is preserved through the normalization layers. 
The generative model is trained with the language modeling loss given by,
\begin{equation}\label{eq:gen_loss}
    \mathcal{L}_G = \mathbb{E}_{x,y}\bigg[-\sum_{i=1}^n \log P_G(x_i|x_{<i}, y)\bigg]
\end{equation}
% L_G= E_(x,z~X) [-∑_(t=1)^n▒log⁡〖P_G (x_t ┤| x_1…x_(t-1),z)〗 ]m
% where $P_G(x_i|x_{<i}, y)$ denotes the probability of generator $G$ predicting the token $x_i$ given all the previous tokens and the control $y$.

We introduce a metric loss as feedback for the degree of metric control achieved during generation. This is defined as the cross-entropy loss between the input target metric and the projected metric for the generated text. The latter is calculated using a fastText \cite{joulin2016fasttext} classifier trained on the outcome on the historical text across various metrics. Such a classifier, which predicts the engagement on held-out test set with high confidence, serves as an indicator of expected engagement on generated text. The metric loss is
\begin{equation}\label{eq:metric_loss}
    \mathcal{L}_{\operatorname{metric}}=\mathbb{E}_{x,y, \tilde{x}=G(x,y)}\big[ -y \log P_F(y|\tilde{x}) \big]
\end{equation}
$P_F(y|\tilde{x})$ denotes the probability of the outcome of the generated text $\tilde{x}$ to be the target metric $y$. We can not directly use this loss in back-propagation because of the discrete sampling of $\tilde{x}$ in the generative model. Thus, we use $P_F(y|\tilde{x})$ as reward and apply REINFORCE algorithm \cite{sutton1999Policy} for policy-gradient based optimization. 

\textbf{Causal-guidance in Generative Model. }\label{section:causalTransformer}
The addition of the target metric as control in input embedding, self-attention mechanism or layer normalization guides the generative model towards the target metric by shifting the language distribution of the generative model. However, an explicit guidance of different aspects of text that influence the outcome metric is absent. To achieve this, we add causal guidance in the generation process. We introduce a causal loss in the above Transformer model to lead the generated text to adopt causally significant features ($t$). The output tokens generated from the Transformer are fed into an SVM that extracts these features from the generated text.
% The model is then optimised on two additional objectives, one, to minimize the mean squared error between the causal features of the input and generated text and the other, to minimize the cross entropy loss between the metric predicted using these features extracted from the output text and the target metric. The causal loss is
% \begin{equation}
% \begin{split}\label{eq:causal_loss}
% \small
%     \mathcal{L}_{\operatorname{causal}} = \mathbf{E}_{x, y, \tilde{x}=G(x,y)} &\big[ (t(x)-t(\tilde{x}))^2 \\
%     - y &\log P_{F'}(y|t(\tilde{x})) \big]
% \end{split}
% \end{equation}
% where $t(x)$ gives the vector of causally significant features extracted from the text $x$ and the $P_{F'}$ is the probability of a fastText model predicting metric as the target metric, given the causal features $t(x)$. Note, this fastText model is trained on causal features extracted on observed data. The MSE ensures that distribution of causal features in generated text is same as that in input and the CE term ensures metric adherence based on the causal features.
The model is then trained with the additional objective of minimizing the cross-entropy loss between the target metric and the predicted outcome metric based on these causal features in output text.
\begin{equation}
\label{eq:causal_loss}
    \mathcal{L}_{\operatorname{causal}} = \mathbb{E}_{x, y, \tilde{x}=G(x,y)} \big[ - y \log P_{F'}(y|t(\tilde{x})) \big]
\end{equation}
where 
% $t$ is the vector of causally significant features in the generated text and the 
$P_{F'}$ is the expected outcome metric given the causal features $t(x)$, estimated using a fastText model trained on causal features extracted from observed data. The proposed causal loss aims at ensuring that the causal features in generated text adheres to target metric by isolating the effect of causal features in text from its context. 
% probability of a fastText model predicting metric as the target metric, given the causal features $t(x)$. This isolates the effect of causal features from the contextual or semantic properties of text. Note, this fastText model is trained on causal features extracted on observed data. The proposed causal loss aims at ensuring that the causal features in generated text adheres to target metric. 
% The model is then optimised on two additional objectives, one, to minimize the cross entropy loss between the lexical/syntactic features of the input and generated text and the other,  to minimize the cross entropy loss between the metric predicted using these features extracted from the output text and the target metric. The causal loss is
% \begin{equation}
% \begin{split}\label{eq:causal_loss}
%     \mathcal{L}_{\operatorname{causal}}=\mathbf{E}_{x, y, \tilde{x}=G(x,y)} & \\
%     \big[ - P_{F'}(y|C(x)) & \log P_{F'}(y|C(\tilde{x})) \big]
% \end{split}
% \end{equation}
% where $C(x)$ gives the vector of causally significant features extracted from the text $x$ and the $P_{F'}$ is the probability of a fastText model predicting metric as the target metric, given the causal features $C(x)$. Thus, the causal loss is equivalent to cross-entropy loss on predicted metric, given causal features for actual text $x$ and generated text $\tilde{x}$.

The resultant loss optimized by the proposed model is a weighted sum of these losses, i.e. $\mathcal{L} = \lambda_G \mathcal{L}_G +\lambda_{\operatorname{metric}} \mathcal{L}_{\operatorname{metric}}+\lambda_{\operatorname{causal}} \mathcal{L}_{\operatorname{causal}}$,
% \begin{equation}
%     \mathcal{L} = \lambda_G \mathcal{L}_G +\lambda_{metric} \mathcal{L}_{metric} + \lambda_{causal} \mathcal{L}_{causal}
% \end{equation}
% L= λ_G L_G+λ_metric L_metric+λ_causal L_causal
where $\lambda_G$, $\lambda_{\operatorname{metric}}$, $\lambda_{\operatorname{causal}}$ are weights for different losses selected by hyper-parameter tuning on validation set.

\subsection{Parallels: Causal CVAE and Transformer}\label{parallel}
In the VAE-based models, we consider the context $c$ and discuss the next sentence ($x$) generation task. At token-level, $c$ is similar to the context $x_{<i}$ in the next token ($x_i$) generation objective. Thus, the decoding term in CVAE loss (first term in Eq. \ref{eq:cvae-non-causal}) is equivalent to $\mathcal{L}_G$ (Eq. \ref{eq:gen_loss}) in the Transformer model. Similarly, the KL divergence between metric prior and posterior distribution in $\mathcal{L}_{V_{nc}}$ (last term in Eq. \ref{eq:cvae-non-causal}) can be equated to the metric loss in Eq. \ref{eq:metric_loss}. The corresponding term in $\mathcal{L}_{V_c}$ (last term in Eq. \ref{eq:cvae-causal}) serves as the causal loss, similar to $\mathcal{L}_{causal}$ in Eq. \ref{eq:causal_loss}. With minor adjustments, this causal guidance framework can be extended to other generative networks in a similar fashion.

%% file: sections/experiments.tex
\begin{comment}
\begin{figure}
\centering
\begin{subfigure}[t]{0.23\textwidth}
    \centering
        \includegraphics[scale=0.28]{pcount_hist.png}
        \caption{Participation Count}
        \label{fig:data_pc}
    \end{subfigure} 
    \begin{subfigure}[t]{0.23\textwidth}
    \centering
        \includegraphics[scale=0.28]{rcount_hist.png}
        \caption{Replies Count}
        \label{fig:data_rc}
    \end{subfigure} 
    \caption{Distribution across metrics on the Webhose Data}
    \label{fig:data-dist-hist}
\end{figure}
\end{comment}

% \begin{table}
% \begin{center}
% \resizebox{.40\textwidth}{!}{
%     \centering
%     \begin{tabular}{l|l|c|c|c}
%     \hline
%         Dataset & Metric & Low & Medium & High \\
%         \hline
%         Webhose & Participation & 20482 & 9181 & 9529 \\
%         (Total:39192)& Replies & 20440 & 9262 & 9490 \\
%         \hline
%         NYT & Comment & 3160 & 3075 & 3168 \\
%         (Total:9403) & Upvote & 3122 & 3126 & 3155 \\
%         \hline
%     \end{tabular}}
%     \caption{Number of samples in across metrics}
%     \label{tab:data-distribution}
%     \end{center}
% \end{table}

\begin{table}
\small
    \centering
    \begin{tabular}{l|l|r|r|r}
    \hline
        Dataset & Metric & Low & Med. & High \\
        \hline
        Webhose & Participation & 20482 & 9181 & 9529 \\
        (Total:39192)& Replies & 20440 & 9262 & 9490 \\
        \hline
        NYT & Comment & 3160 & 3075 & 3168 \\
        (Total:9403) & Upvote & 3122 & 3126 & 3155 \\
        \hline
    \end{tabular}
    \caption{Number of samples in across metrics}
    \label{tab:data-distribution}
\end{table}

%%%%% NEW

\begin{table*}[ht]
\small
    \centering
    \begin{tabular}{c| l c| c cc p{6mm}p{6mm}p{6mm}}
    \toprule
    Metric/ & Model & Variation & Control ($\uparrow$) & Perplexity & BLEURT  & \multicolumn{3}{c}{ROUGE ($\uparrow$)} \\
    
    Dataset & & & \% accuracy& ($\downarrow$) & ($\uparrow$)& 1 & 2 & L \\
    \midrule
    
    % Methods $\downarrow$ & Datasets $\rightarrow$& \multicolumn{7}{|c?}{Webhose}& \multicolumn{7}{c}{NYT}\\
    % \midrule
     \multirow{8}{*}{\rotatebox{90}{\makecell{Participation \\ (Webhose)}}}& \multirow{3}{*}{Transformer} & Baseline GPT-2 & 51.93 & 16.27 & -0.98 & 0.010 & 0.0 & 0.002 \\
    % & $\mathcal{L}_G$ && 54.91 & 17.89 & -0.87 &  0.118 & 0.016 & 0.090& & 49.32 & 23.22 & -0.93 &   0.065 & 0.012 & 0.055\\
     && $\mathcal{L}_G$ & 59.94 & 15.14 & -0.81  & 0.110 & 0.013 & 0.085 \\

    %  & $\mathcal{L}_G$ + $\mathcal{L}_{metric}$ && 61.79 & 3.06 & -0.88  & 0.142 & 0.014 & 0.091 && 52.10 & 15.82 & -0.81 &  0.096 & 0.010 & 0.047 \\

     && $\mathcal{L}_G$ + $\mathcal{L}_{metric}$ & 62.78 & \textbf{3.03} & -0.83  & 0.113 & 0.012 & 0.074 \\

    && Causal model (our) & \textbf{69.86} & 3.19 & -0.79  & \textbf{0.201} & \textbf{0.022} & \textbf{0.130} \\

    % \cline{1-2}\cline{4-9}
    \cline{2-9}
    &\multirow{3}{*}{CVAE} & Baseline CVAE & 51.37 & 34.37 & -0.80  & 0.113 & 0.010 & 0.063 \\

    && metric-guided & 54.43 & 28.21 & -0.69 &  0.179 & 0.017 & 0.099 \\

    && Causal model (our) & 55.66 & 30.03 & \textbf{-0.71} & 0.130 & 0.012 & 0.079 \\

    \bottomrule

    % Replies count
    
    \multirow{8}{*}{\rotatebox{90}{\makecell{Replies\\(Webhose)} }}& \multirow{3}{*}{Transformer} & Baseline GPT-2 & 51.79 & 17.76 & -0.91 &  0.005 & 0.0 & 0.005 \\

    %  & $\mathcal{L}_G$ && 55.40 & 23.53 & -0.89 &  0.093 & 0.009 & 0.077 && 45.37 & 23.69 & -0.91 &  0.071 & 0.026 & 0.064 \\

     && $\mathcal{L}_G$ & 59.87 & 13.94 & -0.85 &  0.051 & 0.004 & 0.043 \\

    % & $\mathcal{L}_G$ + $\mathcal{L}_{metric}$ && 61.18 & \textbf{2.99} & -0.86 &  0.142 & 0.014 & 0.091   &&  51.12 & 15.70 & -0.85  & 0.120 & 0.014 & 0.047 \\

     && $\mathcal{L}_G$ + $\mathcal{L}_{metric}$   & 60.17 & 3.48 & -0.79 &  0.107 & 0.011 & 0.070 \\
    && Causal model (our)& \textbf{68.27} & \textbf{3.12} & -0.81  & \textbf{0.211} & \textbf{0.022} & \textbf{0.133} \\

    \cline{2-9}
    &\multirow{3}{*}{CVAE} & Baseline CVAE & 50.58 & 38.41& -0.89 & 0.046 & 0.001 & 0.035 \\

    && metric-guided & 56.14 & 20.58 & -0.8 & 0.124 & 0.002 & 0.072 \\

     && Causal model (our) & 60.00 & 30.24 & \textbf{-0.76} &  0.031 & 0.001 & 0.022\\
    
    \bottomrule
    %% NEW
    
     \multirow{8}{*}{\rotatebox{90}{\makecell{Comments \\ (NYT)}}}& \multirow{3}{*}{Transformer} & Baseline GPT-2 & 37.24 & 27.45 & -0.83 & \textbf{0.140} & \textbf{0.088} & \textbf{0.135}\\
    % & $\mathcal{L}_G$ && 54.91 & 17.89 & -0.87 &  0.118 & 0.016 & 0.090& & 49.32 & 23.22 & -0.93 &   0.065 & 0.012 & 0.055\\
     && $\mathcal{L}_G$ & 49.85 & 23.59 & -0.87 &  0.095 & 0.051 & 0.088\\

    %  & $\mathcal{L}_G$ + $\mathcal{L}_{metric}$ && 61.79 & 3.06 & -0.88  & 0.142 & 0.014 & 0.091 && 52.10 & 15.82 & -0.81 &  0.096 & 0.010 & 0.047 \\

     && $\mathcal{L}_G$ + $\mathcal{L}_{metric}$ & 53.82 & 14.99 & -0.89 &  0.10 & 0.011 & 0.052 \\

    && Causal model (our) & \textbf{54.36} & \textbf{13.18} & \textbf{-0.81}  & 0.10 & 0.01 & 0.049 \\

    \cline{2-9}
     &\multirow{3}{*}{CVAE} &Baseline CVAE &  39.12 & 58.35 & -1.41 &  0.059 & 0.002 & 0.031\\

    && metric-guided &   44.42 & 41.64 & -1.32 &  0.069 & 0.003 & 0.036 \\

    && Causal model (our) & 54.59 & 40.02 & -1.29 &  0.064 & 0.003 & 0.032\\

    \bottomrule
    
    \multirow{8}{*}{\rotatebox{90}{\makecell{Upvotes \\ (NYT)} }} & \multirow{3}{*}{Transformer} & Baseline GPT-2 & 39.49 & 27.44 & -0.83 &  \textbf{0.132} & \textbf{0.080} & \textbf{0.127}\\

    %  & $\mathcal{L}_G$ && 55.40 & 23.53 & -0.89 &  0.093 & 0.009 & 0.077 && 45.37 & 23.69 & -0.91 &  0.071 & 0.026 & 0.064 \\

     && $\mathcal{L}_G$ &  46.02 & 23.57 & -0.88  & 0.077 & 0.032 & 0.070\\

    % & $\mathcal{L}_G$ + $\mathcal{L}_{metric}$ && 61.18 & \textbf{2.99} & -0.86 &  0.142 & 0.014 & 0.091   &&  51.12 & 15.70 & -0.85  & 0.120 & 0.014 & 0.047 \\

     && $\mathcal{L}_G$ + $\mathcal{L}_{metric}$   & 53.66 & 14.93 & -0.82 &  0.110 & 0.011 & 0.053 \\
    && Causal model (our) & \textbf{59.54} & \textbf{13.19} & \textbf{-0.80}  & 0.103 & 0.010 & 0.051\\

    \cline{2-9}
    &\multirow{3}{*}{CVAE} & Baseline CVAE & 37.06 & 72.68 &  -0.89  & 0.057 & 0.002 & 0.031\\

    && metric-guided & 43.21 & 65.94 & -0.84  & 0.064 & 0.002 & 0.036\\

     && Causal model (our) & 53.96 & 57.70 & -0.84  & 0.056 & 0.001 & 0.030\\
    
    \bottomrule
    
    \end{tabular}
    \caption{Automatic Evaluation for Webhose (Participation, Reply count) and NYT (Comments, Upvotes) Datasets. The causal Transformer model beats all other methods on metric control while achieving comparable fluency.}
    \label{tab:results-quantitative-webhose}
\end{table*}

%%%%% END NEW

\section{Experiments}
\subsection{Datasets}
% \noindent\textbf{Datasets.} 
We experiment with $2$ text datasets: NYT comments, which comprises articles with comments and metrics such as upvote and comments count and the Webhose\footnote{\url{https://webhose.io/free-datasets/news-articles-that-mention-corona-virus/}} dataset comprising of articles and comments with metrics such as total participation on articles, replies count, and various social media reactions for these articles. These metrics are used as target goal for article text generation. We filter and pre-process\footnote{Preprocessing details in Appendix C} this data resulting in 39k article data which we use for our training with a train-dev-test split of $80$-$10$-$10$ (Table \ref{tab:data-distribution}). We categorize the target metrics into high, medium, and low classes, resulting in categorical target goal (e.g., high/ low replies count). 

\subsection{Training}
% \noindent\textbf{Training details.} 
For causal model, we use two sequential feed forward neural networks with $5$ dense layers of size $128$, each followed by an activation layer, for the treatment and potential outcome network trained with Adam optimizer \cite{kingma2014adam}. The parts of speech (POS) are extracted using the POS tagging in textblob\footnote{\url{https://textblob.readthedocs.io}} library. Both treatment and potential outcome networks are trained on $90$-$10$ train-test split over $10$ epochs.

For CVAE, we use a bidirectional recurrent neural network (bi-RNN), which encodes each context sentence to a fixed $300$-sized vector. We pass these vectors through another GRU network with one hidden-layer of $600$-dimension, resulting in the context vector $c$. The decoder network is also a one-layer GRU with dimensionality $400$. The end-to-end model is trained with an Adam optimizer.

We use a Transformer model with $16$ multi-attention heads with latent dimension of $768$ and a vocabulary size of $50527$ with BPE encoding \cite{sennrich-etal-2016-neural}. We use the GPT-2 \cite{radford-wu-2018-gpt-2} model with $117M$ parameters pre-trained on the WebText dataset to initialize our model and then fine-tune it with NYT and Webhose datasets using our causal metric-guided framework. For causal variants, the causal vector $t$ is extracted from the generated text based on a pre-determined list of causally significant features (identified beforehand using ATE analysis in section \ref{section:causal-effect}).

\begin{figure*}
    \centering
    \begin{subfigure}[t]{0.2\textwidth}
    \centering
        \includegraphics[scale=0.24]{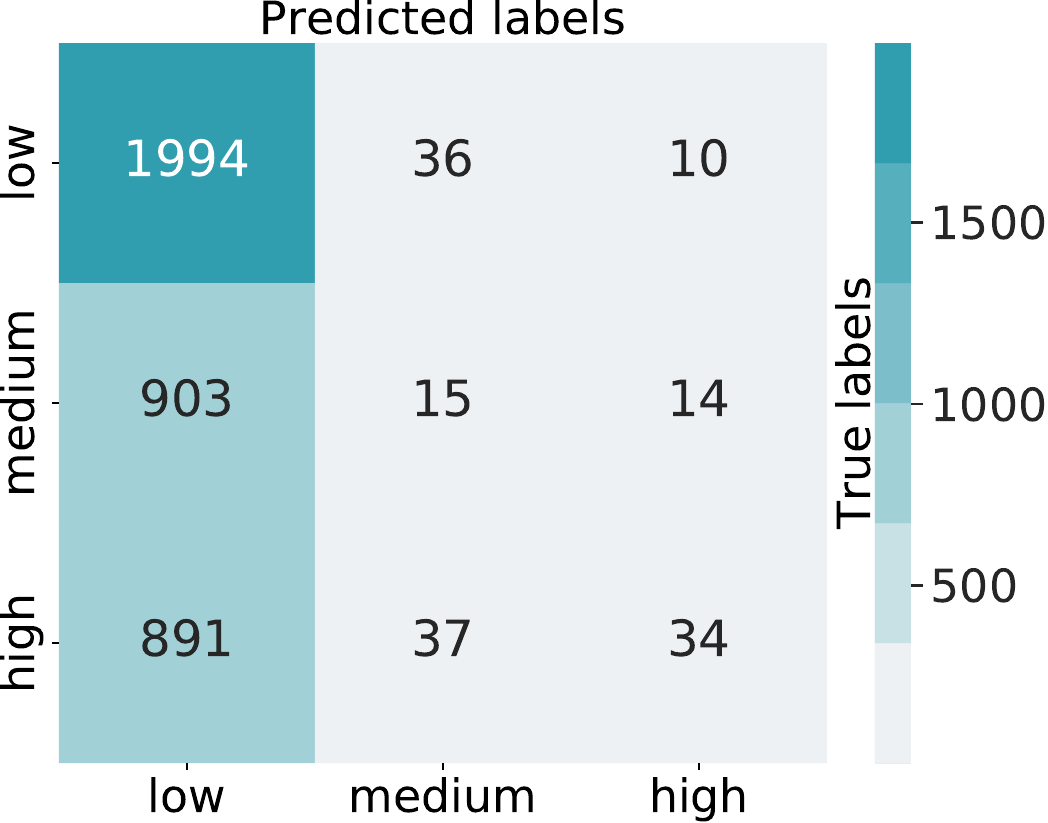}
        \caption{Baseline GPT2}
        \label{fig:baseline}
    \end{subfigure}
    \begin{subfigure}[t]{0.2\textwidth}
    \centering
        \includegraphics[scale=0.24]{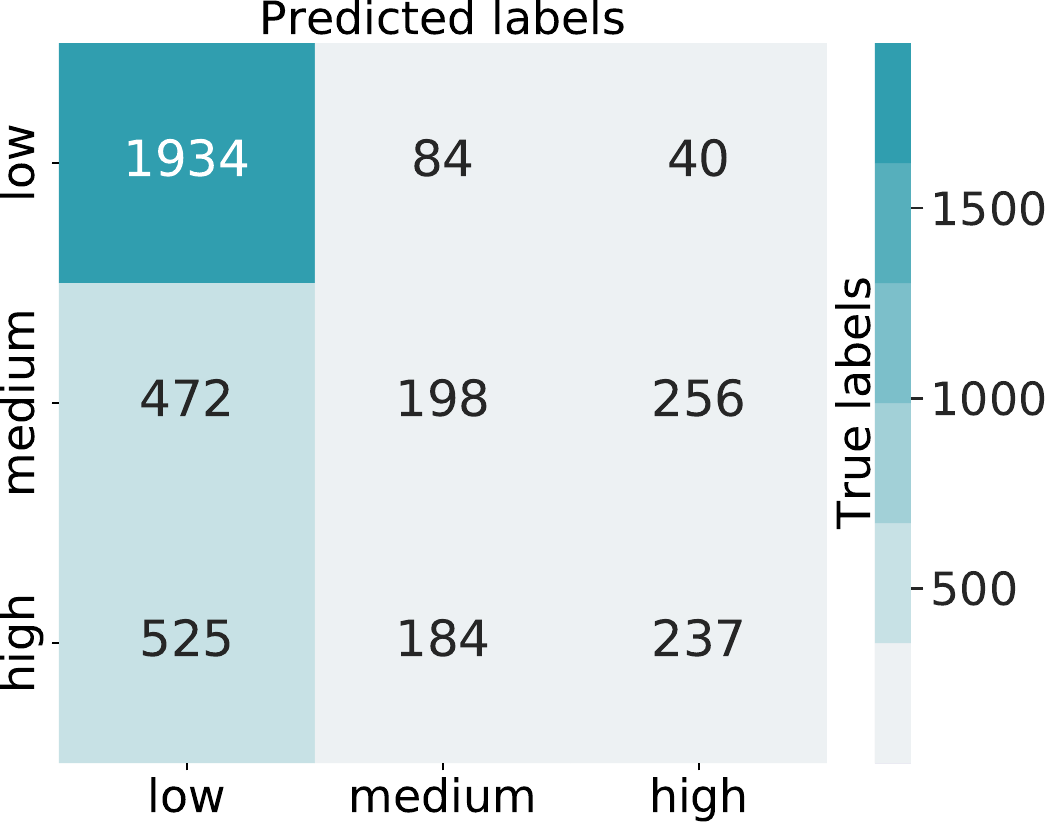}
        \caption{$\mathcal{L}_G$}
        \label{fig:neither}
    \end{subfigure}
    \begin{subfigure}[t]{0.2\textwidth}
    \centering
      \includegraphics[scale=0.24]{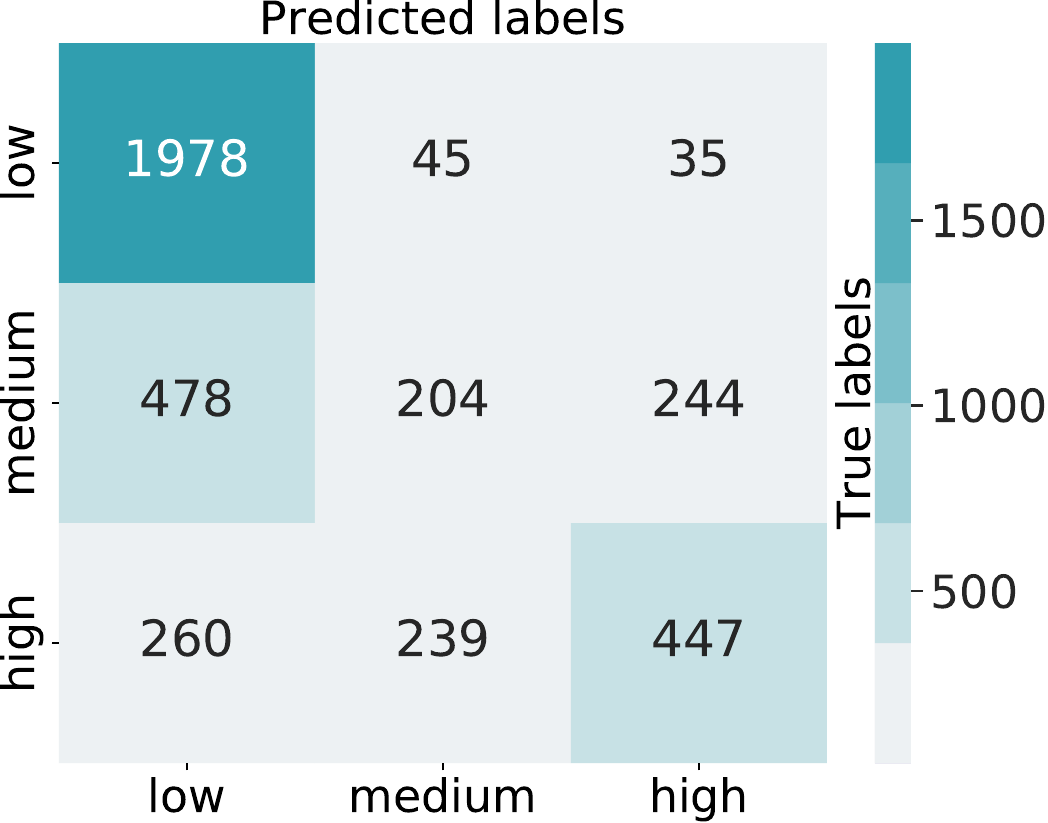}
      \caption{$\mathcal{L}_G+\mathcal{L}_{metric}$}
      \label{fig:metric}
    \end{subfigure}
    %   \begin{subfigure}[t]{0.18\textwidth}
    % \centering
    %     \includegraphics[scale=0.25]{Images/confusion_matrix/gpt2_topic_metric_pcount.png}
    %     \caption{$\mathcal{L}_G+\mathcal{L}_T+\mathcal{L}_{metric}$}
    %     \label{fig:metricTopic}
    % \end{subfigure}
    \begin{subfigure}[t]{0.24\textwidth}
    \centering
        \includegraphics[scale=0.24]{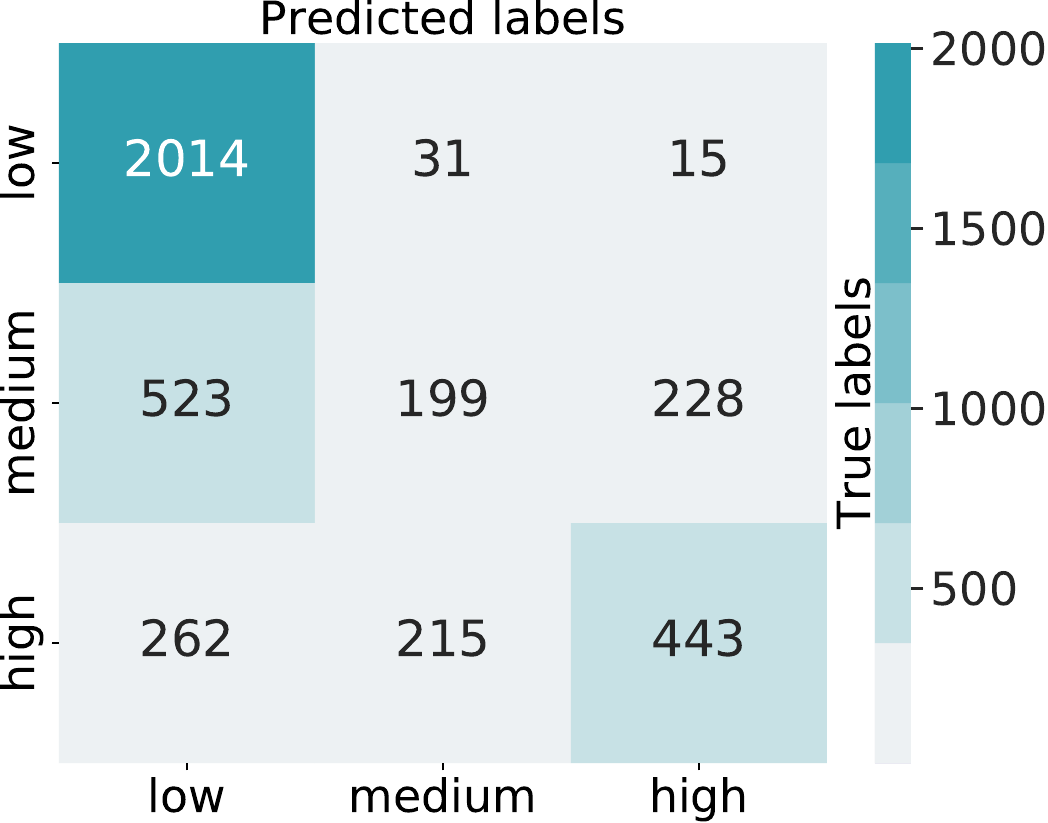}
        \caption{Causal }
        \label{fig:causal}
    \end{subfigure}
    
    \caption{Class-wise performance for Transformer-based model variants.}
    \label{fig:confusion_matrix}
\end{figure*}

\subsection{Evaluation metrics}
%We measure the generated outputs from the model across the following automatic metrics:
%\begin{enumerate}[leftmargin=*]
    \textbf{Control}: 
    % We evaluate whether the generated text is on-target in terms of the metric specified. 
    We measure target control accuracy against predicted outcome metric in the generated text using fastText classifiers trained on available data. The classifiers have test accuracy of $79.8\%$, $81.4\%$, $80\%$ and $79.9\%$ for participation, replies, comment, upvotes counts, respectively.\\ %Percentage accuracy is reported.\\ %We evaluate the generated output against these classifiers and report the percentage accuracies.
    % \underline{Fluency}: We measure the text fluency and the language model quality across following metrics.\\
    % (i) Perplexity: Perplexity is a measure of likelihood of the generated sentence on a language model. We use a pre-trained GPT-2 model to evaluate perplexity of generated text. A lower value is preferred. \\%indicates that the generated language fits with the actual test content in the corpus.\\
    % (ii) ROUGE\cite{lin-2004-rouge}: We measure how well the generated article relates to the reference text from the same keywords and target metric values.\\
    % (iii) BLEURT: BLEURT \cite{sellam-etal-2020-bleurt} is a pre-trained evaluation metric based on BERT \cite{devlin-etal-2019-bert} that provides a robust measure for reference-based text generation. Here, the reference is taken to be the articles having the same keywords and target control as the test data. \\
    \textbf{Fluency}: We measure the text fluency and the language model quality using perplexity, ROGUE \cite{lin-2004-rouge} and BLEURT \cite{sellam-etal-2020-bleurt} scores. The perplexity is a measure of likelihood of the generated sentence on a language model. We use a pre-trained GPT-2 model to evaluate text perplexity. A lower value is preferred. BLEURT is a pre-trained evaluation metric based on BERT \cite{devlin-etal-2019-bert} that provides a robust measure for reference-based text generation. We calculate ROGUE and BLEURT scores against reference articles in test data with same keywords and target.
    % BLEURT shows the ability to generalise well and correlates well with human judgements.\\
    %\\Readability Scores: We use Coleman-Liau index\footnote{https://github.com/shivam5992/textstat} that estimates the understandability of the text in terms of the U.S. grade level thought necessary to comprehend the text. 
    % \textbf{Human-judgements}: To validate whether the generated text shows a substantial change beyond empirical metrics, we conduct an experiment with crowdsourced annotators on Amazon Mechanical Turk. The annotators are asked to independently rate baseline, causal, and non-causal variants for whether they would like/share the content, simulating participation count, on a 5 point likert scale. To gauge preference among generated variants, a pairwise comparison of baseline and non-causal outputs with the causal output is also gathered\footnote{10 annotations per sample}. %Variants with the same target metric value and topic prompts are used for uniform comparison.

%\end{enumerate}
\section{Results}\label{section:results}
We compare causal and non-causal variants of the proposed CVAE and Transformer-based models. %and \ref{tab:results-quantitative-nyt} for Webhose and NYT datasets respectively.
In the Transformer variants, we evaluate the performance with metric added as a guide in embedding, attention, and normalization layers, trained with {$\mathcal{L}_G$} (Eq. \ref{eq:gen_loss}). Next, we introduce the metric loss to add feedback for adherence to target metric, training the model with $\mathcal{L}_G$ + $\mathcal{L}_{\operatorname{metric}}$ (Eq. \ref{eq:metric_loss}).
% The non-causal model is trained with $\mathcal{L}_G$ + $\mathcal{L}_{\operatorname{metric}}$ and 
The final proposed causal model is trained with $\mathcal{L}_G$ + $\mathcal{L}_{\operatorname{metric}}$ + $\mathcal{L}_{\operatorname{causal}}$ (Eq. \ref{eq:causal_loss}).
% (eq. \ref{eq:gen_loss}, \ref{eq:metric_loss}, \ref{eq:causal_loss}, \ref{eq:topic_loss}). 
For CVAE, non-causal and causal models are trained with $\mathcal{L}_{V_{nc}}$ and $\mathcal{L}_{V_c}$ (Eq. \ref{eq:cvae-non-causal}, \ref{eq:cvae-causal}) respectively.
We fine-tune a GPT-2 \cite{radford-wu-2018-gpt-2} model with metric token added to the prompt for control, similar to \cite{keskar2019ctrl}, and use it as a baseline. %, but not explicitly conditioned for. %Such a baseline helps establish if large pre-trained language models are able to accomplish fine-grained control over the generated text, in terms of target metric, without explicitly modelling this control or incorporating the causal insights of various features in text that potentially leads to high engagement.
We also use the method proposed by \cite{zhao-etal-2017-kgcvae} as the baseline CVAE model.

\begin{table}[t]
    \centering
    \begin{subtable}{0.45\textwidth}
    \centering
    \begin{tabular}{lcc}
    Treatment & Loss & Accuracy \\
    \hline
    Word Count & 0.1791 & 0.9301 \\
    Sent Count & 0.2268 & 0.9266 \\
    Noun Count & 0.1520 & 0.9520 \\
    Verb Count & 0.1437 & 0.9592 \\
    Adjective Count & 0.2133 & 0.9349 \\
    Adverb Count & 0.1863 & 0.9431 \\
    Pronoun Count & 0.1522 & 0.9377 \\
    \hline
    \end{tabular}\label{tab:propensity-ce}
    \caption{Propensity scoring model}
    \end{subtable}\\
    % \vspace{0.3cm}
    \begin{subtable}{0.45\textwidth}
    \centering
    \begin{tabular}{lcc}
    Outcome metrics & MAE & Accuracy \\
    \hline
    Upvotes & 0.1357 & 0.9157 \\
    Replies count & 0.2359 & 0.8455 \\
    Discussion depth & 0.2549 & 0.8322 \\
    Comment count & 0.1438 & 0.9104 \\
    \hline
    \end{tabular}\label{tab:potential-outcome-mae}
    \caption{Potential outcome model}
    \end{subtable}
    \caption{Loss and test accuracy of of causal effect identification models}
    \label{tab:causal-accuracy}
\end{table}

\begin{figure*}[t]
\centering
  \begin{subfigure}[t]{0.4\textwidth}
        \includegraphics[width=0.95\linewidth]{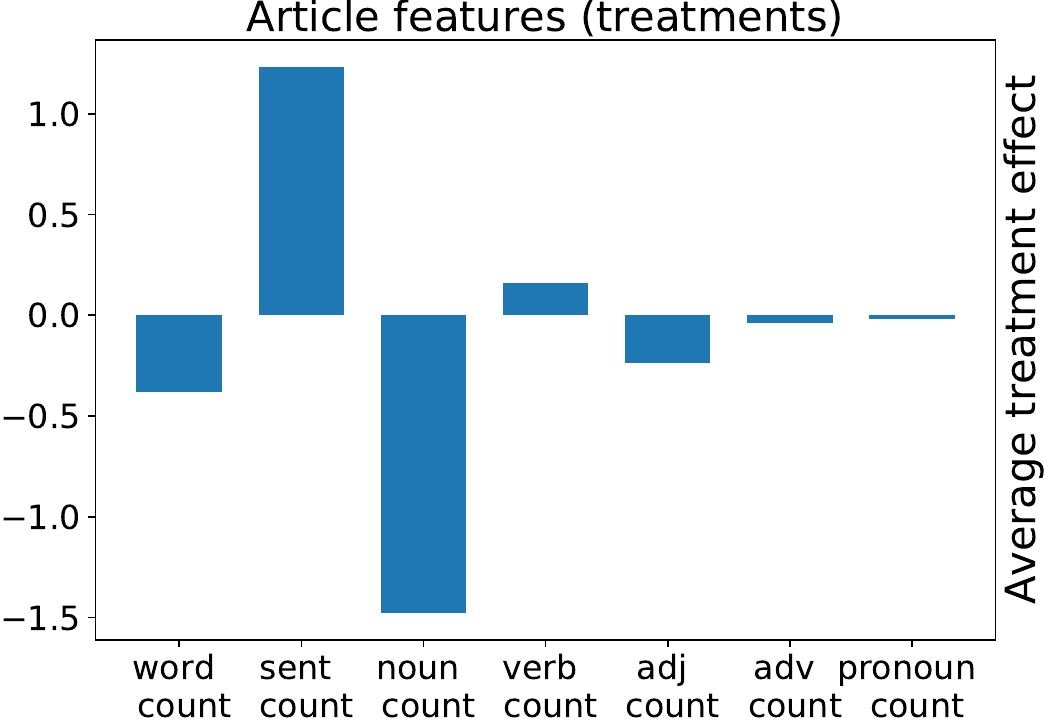}
        \caption{Participation Count}
        \label{fig:ate_recomm}
    \end{subfigure}
    % \hfill
    \begin{subfigure}[t]{0.4\textwidth}
        \includegraphics[width=0.95\linewidth]{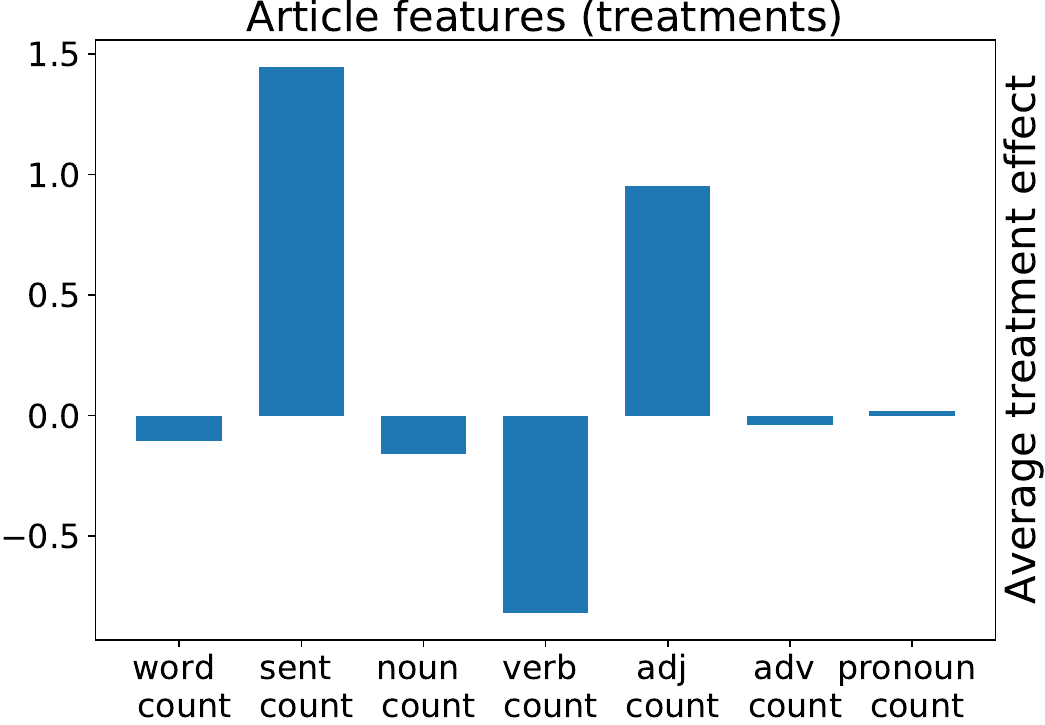}
        \caption{Replies Count}
        \label{fig:ate_comm}
    \end{subfigure}
    \caption{Average treatment effect of features like word count, sentence count, POS tag counts across metrics.}
    \label{fig:ate}
\end{figure*}

As seen in Table \ref{tab:results-quantitative-webhose}, adding metric as explicit guide improves accuracy both in Transformer and CVAE models, and the causal models outperforms all other variants in the same architecture. Additionally, our variants are at par in text quality, with the Transformer models performing notably better on language fluency than CVAE models. We attribute this to generative pre-training with large corpus equipping Transformer-based language model with fluent language generation. Note that, given the free-form nature of generative task, the references considered for ROUGE and BLEURT are a poor fit as the generation space could be pretty large. This is reflected in low scores for these metrics across all models. Hence, low perplexities are a better indication of generation fluency.

Causal CVAE exhibits better metric control than the non-causal and baseline CVAE but performs poorer than the causal Transformer model. This could also be an artifact of language quality since the underlying classifiers are trained on fluent language. 
Across Transformer variations, addition of metric loss and causal guidance improves metric control, validating our hypothesis. It is interesting to note that the perplexity drops substantially on adding the metric loss in Transformer-based model. This could raise the question on how additional losses (constraints) could result in more fluent generation. We emphasize that, in baseline and all other variants, the constraint is on the target metric. Thus, both baseline GPT-2 and modified Transformer (with only $\mathcal{L}_G$) attempt to align their generation space to this target. An inadequate alignment of generation space to the desired control is likely to result in noisy generations. In that sense, metric/causal do not add more constraints, rather add feedback to meet the specified constraint (goal), leading to more controlled and less noisy generations. This would potentially explain higher perplexities observed in the first two variants. 
% This could be one possible reason behind higher perplexities observed in these two variations. Intuitively, the metric/causal losses do not simply add more constraints, rather add sufficient feedback for already present constraint. This could imply that adding metric loss not only improves control, but also reduces noisy generations.\\
% \textbf{Human-judgements.} The human study included 300 participants, with over 5 annotations per example. In the independent scoring task, while we do not observe significant preference across models, causal model output is marginally preferred over others in terms of text quality, across classes. (model:[high,low,medium]; baseline:[3.57,3.55.3.53]; non-causal:[3.66.3.57,3.54]; causal:[3.64,3.65,3.7]). For the comparative task, the causal model is preferred in 53\% cases over non-causal and 55.68\% over the baseline model. With `high' participation count as target metric, causal model trumps 60\% over the baseline, hence further establishing confidence in this work.

\noindent\textbf{Class-wise Performance.} Table \ref{tab:results-quantitative-webhose} aggregates results across target classes. To compare the performance across high/medium/low class, we record class-wise metric accuracy. Fig. \ref{fig:confusion_matrix} shows confusion matrices for Transformer-based variants with high/medium/low participation count as target. Across methods, we observe that controlling for medium target metric is harder than either of the other classes. Compared to the baseline, variants with causal guidance and metric loss show improved performance for both high and low target class. Our proposed causally guided Transformer model is the best performing model on per class-level as well, confirming the efficacy of our proposed approach across different target classes.

\noindent\textbf{Causal Feature Identification. }
%Since we do not observe the outcome on hypothetical treatment (\textit{counterfactual}), we can not directly evaluate the ATE estimates described in section \ref{section:causal-effect}. Instead 
Table \ref{tab:causal-accuracy} shows the accuracy of the propensity scoring and potential outcome models. Our propensity scoring models have accuracy $>0.92$ for all treatment features and the potential outcome model performs well for \textit{Upvote} and \textit{Comment count}. We use these as target metrics in generative models for NYT dataset. Similar analysis on Webhose data yields \textit{Participation} and \textit{Replies count} as target metric. Fig. \ref{fig:ate} shows Average Treatment Effect (ATE) of various text features on these outcome metrics. We empirically choose significance level of $0.1$ and consider features with ATE of greater than $0.1$ (in magnitude) as `causally significant' features. We include these as causal features in the generative models. 

%pre-final
% \begin{figure}[t]
%   \begin{subfigure}[t]{0.218\textwidth}
%     \centering
%         \includegraphics[scale=0.22]{figures/gen_feature_analysis_causal_pcount.pdf}
%         \caption{Causal Model}
%         \label{fig:gen_causal}
%     \end{subfigure}
%     \begin{subfigure}[t]{0.24\textwidth}
%     \centering
%         \includegraphics[scale=0.22]{figures/gen_feature_analysis_baseline_pcount.pdf}
%         \caption{Baseline Model}
%         \label{fig:gen_baseline}
%     \end{subfigure}
%     \caption{Comparison of textual features in text generated by causal vs baseline Transformer model}
%     \label{fig:gen_feat_analysis}
% \end{figure}

%final
\begin{figure*}[t]
\centering
  \begin{subfigure}[t]{0.43\textwidth}
    \includegraphics[width=0.95\linewidth]{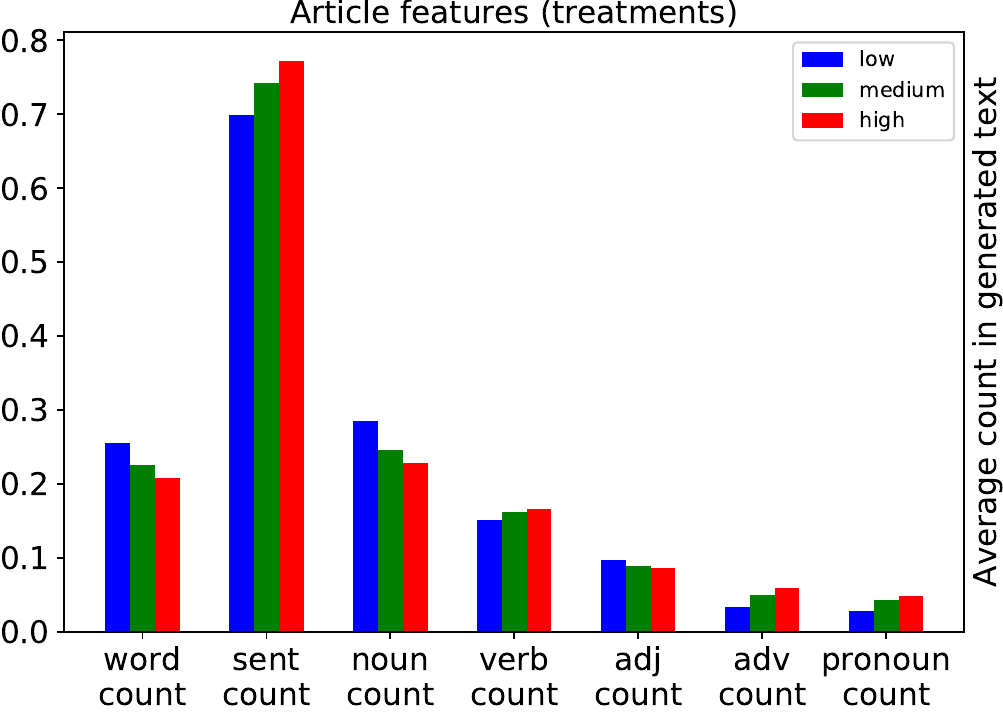}
    \caption{Causal Model}
    \label{fig:gen_causal}
    \end{subfigure}
    \begin{subfigure}[t]{0.43\textwidth}
    \includegraphics[width=0.95\linewidth]{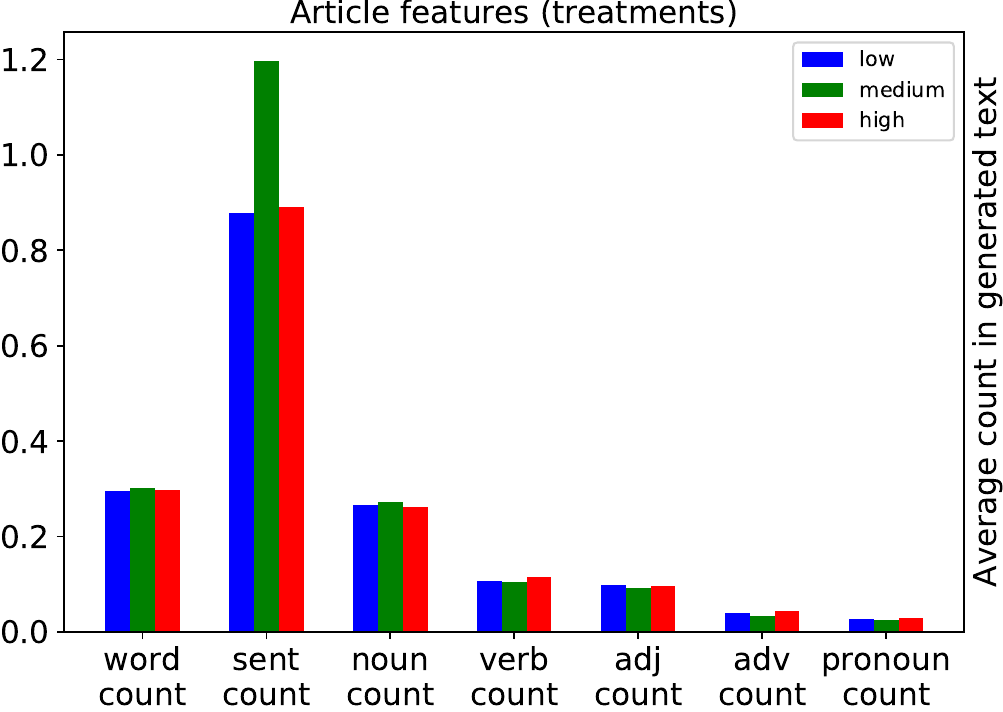}
    \caption{Baseline Model}
    \label{fig:gen_baseline}
    \end{subfigure}
    \caption{Comparison of textual features in text generated by causal vs baseline Transformer model}
    \label{fig:gen_feat_analysis}
\end{figure*}

\noindent\textbf{Causal Analysis. }
We note that the fastText classifiers used for metric evaluation have relatively low accuracy (although much better than a random $33\%$ classification). We attribute this to high variability in the text and unpredictability of resulting engagement. As discussed previously, a causal analysis of historical text accounts for semantic and topical variation. Similarly, a causal analysis of generated data, and subsequent comparison with historical trends, could compensate for any potential inadequacies of classifier-based evaluation. 
To this end, we perform a causal analysis of the text generated by the baseline and our proposed model. 

We generate text with high, medium and low target participation count (\emph{pcount}) as target and record average value of various treatment features (Fig. \ref{fig:gen_feat_analysis}). Here, the word and sentence counts are normalized and POS features are fraction of words with certain POS tag over total number of words in the generated text. We test the adoption of `causally significant' features in the causal model by analyzing feature distributions of text generated by causal model and baseline Transformer model across classes (high/medium/low). For instance, word count has a negative ATE on \textit{pcount} (Fig. \ref{fig:ate_recomm}). Thus, we would expect a text with higher word count to have lesser pcount. As seen in Fig. \ref{fig:gen_causal}, our causal model with `high' target \textit{pcount} generated articles with lower word count on average than the causal model with `low' target (red and blue bars in first group in Fig. \ref{fig:gen_causal} respectively). Similar trends are observed across other `causally significant' treatment features. In contrast, the text generated by baseline model (Fig. \ref{fig:gen_baseline}) either do not show significant variation in these features across text generated with high, medium and low target or the difference is inconsistent, reflecting the lack of control over aspects of text in baseline models where generation is only guided by target metric. %Note that the features like adverb count and pronoun count that have very low ATE are not included for guidance in our causal models. Consequently the generated text do not follow the trends dictated by these features.
As these features, by definition, significantly impact the outcome; this analysis adds further confidence in stronger adherence to the target metric in our proposed causal approach over the baseline.

%% file: sections/conclusion.tex
\section{Conclusion}
We present a framework for causally aware metric-guided generation in VAE and Transformer-based models. We successfully identify causally significant text features using causal analysis and incorporate them into the generative model. We show that integrating causal guidance in guided generation enables better control over the target metric, while maintaining language quality. Our proposed causally guided Transformer model shows improved performance across datasets. Moreover, we show that the generated text adheres to these causal features, in line with their observed effect in historic data. 
% This work is an attempt to connect the textual causal inference literature to the work on controlled generation.
This exploration opens up avenues for leveraging causality for controlled generation. %One possible extension is combining a VAE with pre-trained Transformer models as encoder and decoder and then training the entire network together with the proposed causal CVAE. %To the best of our knowledge, our work is the first attempt in integrating causal insights into generation and opens up an avenue for exploiting such insights for more controlled generation. One possible extension is combining a VAE with pre-trained Transformer models as encoder and decoder and then training the entire network together with the proposed causal CVAE.

%% file: sections/ethics.tex
\paragraph{Ethics Statement.} We recognize and acknowledge that our work carries a possibility of misuse for fake news generation, the same as any text generation system. We strongly recommend coupling any such technology with a fake news detection and review system before deployment. We do not believe that our method exacerbates fake news generation as it aims to optimize syntactic and surface-level features, and not topical or semantic features. On the contrary, having a causal guidance towards these specific factors may guide models to focus on these features and deter them from other non-desirable optimization of content. The data and approaches for generating text that optimizes for clicks exist already. Our proposed approach adds a nuanced control on the linguistic features to optimize for generating desirable content, rather than unconstrained optimization for clicks.

%% file: sections/appendix.tex
% \begin{figure}[t]
%     \centering
%     \includegraphics[width=0.3\textwidth]{figures/causal_graph_1.pdf}
%     \caption{VAE Graph for conditional generation}
%     \label{fig:non-causal-graph}
% \end{figure}

\section{Conditional Variational Autoencoder}
\subsection{Non-Causal CVAE}
\label{apx:CVAE}
The graph for non-causal conditional generation using variational autoencoder is shown in Fig. \ref{fig:causal-graph} (left). As discussed in section 4.1, we approximate the intractable posterior distribution $p_\theta(z|x,c,y)$ with the recognition network $q_\phi(z|x,c,y)$, where
\begin{equation}
\label{eq:q-phi}
    q_\phi(z|x,c,y)=q_{\phi}(z,y|x,c)q_{\phi}(y|x,c)
\end{equation}
The variational parameters $\phi$ are chosen such that the approximate posterior distribution $q_\phi(z|x,c,y)$ is as close to the true posterior distribution $p_\theta(z|x,c,y)$ as possible. This is done by minimizing the KL divergence between the two distributions. Thus,
\begin{equation}
    \phi^{*} = \underset{\phi}{\operatorname{argmin}} \operatorname{KL}[q_\phi(z,y|x,c)||p_\theta(z,y|x,c)],
\end{equation}
where the KL divergence is given by,
\begin{equation}
\label{eq:KL}
\begin{split}
    \operatorname{KL}[q_\phi&(z,y|x,c)||p_\theta(z,y|x,c) \\
    &= \mathbf{E}_{q_\phi(z,y|x,c)}\bigg[\log \frac{q_\phi(z,y|x,c)}{p_\theta(z,y|x,c)}\bigg] \\
    &= \mathbf{E}_{q_\phi(z,y|x,c)}\bigg[\log q_\phi(z,y|x,c) \\
    &- \log \frac{p_{\theta}(x,c,z,y)}{p_{\theta}(x|c)}\bigg].
\end{split}
\end{equation}
Rearranging equation \ref{eq:KL} gives,
\begin{equation}
\label{eq:px}
\begin{split}
    \log p_{\theta}(x) &= \operatorname{KL}[q_\phi(z,y|x,c)||p_\theta(z,y|x,c) \\
    &+ \mathbf{E}_{q_\phi(z,y|x,c)} \big[\log p_{\theta}(x,c,z,y)\\
    &-\log q_{\phi}(z,y|x,c) \big]
\end{split}
\end{equation}

We want to minimize the KL divergence term on R.H.S. of equation \ref{eq:px}. Since, the KL divergence is $\geq 0$, the variational lower bound on the log likelihood $\log p_{\theta}(x)$ is given by 
\begin{equation}
\label{eq:l_nc}
\begin{split}
    \mathcal{L}(\theta, \phi;&x,c,y) = \mathbf{E}_{q_\phi(z,y|x,c)} \big[\log p_{\theta}(x,c,z,y) \\
    &- \log q_{\phi}(z,y|x,c) \big]\\
    &= \mathbf{E}_{q_\phi(z,y|x,c)} \big[\log [p_{\theta}(x|c,z,y) p(z,y|c)] \\
    &- \log q_{\phi}(z,y|x,c)\big] \\
    &= \mathbf{E}_{q_\phi(z,y|x,c)} \log p_{\theta}(x|c,z,y) \\ 
    &- \operatorname{KL}\big[q_{\phi}(z,y|x,c)||p_{\theta}(z,y|c)\big]
\end{split}
\end{equation}
Using equation \ref{eq:q-phi}, we get
\begin{equation}
\begin{split}
    &\operatorname{KL}\big[q_{\phi}(z,y|x,c)||p_{\theta}(z,y|c)\big]\\
    &= \mathbf{E}_{q_\phi(y|x,c)} \operatorname{KL} \big[q_{\phi}(z|x,c,y)||p_{\theta}(z|c,y) \big] \\
    &+ \operatorname{KL} \big[q_{\phi}(y|x,c)||p_{\theta}(y|c) \big]
\end{split}
\end{equation}
Replacing in equation \ref{eq:l_nc}, we get the variational lower bound for non-causal CVAE as 
\begin{equation}
\begin{split}
    &\mathcal{L}(\theta, \phi;x,c,y) =  \mathbf{E}_{q_\phi(z,y|x,c)} \log p_{\theta}(x|c,z,y) \\ 
    &- \mathbf{E}_{q_\phi(y|x,c)} \operatorname{KL} \big[q_{\phi}(z|x,c,y)||p_{\theta}(z|c,y) \big] \\
    &- \operatorname{KL} \big[q_{\phi}(y|x,c)||p_{\theta}(y|c) \big]
\end{split}
\end{equation}

% \begin{equation}
% \begin{split}
%     \operatorname{KL}[q_\phi&(z,y|x,c)||p_\theta(z,y|x,c) \\
%     &= \mathbf{E}_{q_\phi(z,y|x,c)}\bigg[\log \frac{q_\phi(z,y|x,c)}{p_\theta(z,y|x,c)}\bigg] \\
%     &= \mathbf{E}_{q_\phi(z,y|x,c)}\bigg[\log q_\phi(z,y|x,c) \\
%     &- \log \frac{p_{\theta}(x|z,y,c)p(z,y|c)}{p_{\theta}(x)}\bigg] \\
%     &= \mathbf{E}_{q_\phi(z,y|x,c)}\big[\log p_{\theta}(x|z,y,c) \big] \\
%     &- \operatorname{KL}\big[q_{\phi}(z,y|x,c)||p_{\theta}(z,y|c) \big]
% \end{split}
% \end{equation}

\subsection{Causal CVAE}

As discussed in section 4.2, we add causal guidance in CVAE framework by adding the treatment vector $t$ for aligning the latent space of the Variational Autoencoder. The posterior distribution for the causal-CVAE graph in Fig. \ref{fig:causal-graph} (right) is approximated by $q_{\phi}(z|x,c,y)$. Similar to equation \ref{eq:l_nc}, we get the variational lower bound for causal CVAE as 
\begin{equation}
\label{eq:l_c}
\begin{split}
    \mathcal{L}(\theta, \phi;t,x,c,y) &= \mathbf{E}_{q_\phi(z,y|t,x,c)} \big[\log p_{\theta}(t,x,c,z,y) \\
    &- \log q_{\phi}(z,y|t,x,c) \big]\\
    &= \mathbf{E}_{q_\phi(z,y|t,x,c)} \big[\log [p_{\theta}(t|x,c,z,y) \\ 
    & p_{\theta}(x|c,z,y) p(z,y|c)] \\
    &- \log q_{\phi}(z,y|t,x,c)\big] \\
    &= \mathbf{E}_{q_\phi(z,y|t,x,c)} \log p_{\theta}(t|x,c,z,y) \\ 
    &+ \mathbf{E}_{q_\phi(z,y|t,x,c)} \log p_{\theta}(x|c,z,y) \\ 
    &- \operatorname{KL}\big[q_{\phi}(z,y|t,x,c)||p_{\theta}(z,y|c)\big].
\end{split}
\end{equation}
The conditional posterior $q_{\phi}(z,y|t,x,c)$ is given by
\begin{equation}
    q_\phi(z|t,x,c,y)=q_{\phi}(z,y|t,x,c)q_{\phi}(y|t,x,c).
\end{equation}
Thus,
\begin{equation}
\begin{split}
    &\operatorname{KL}\big[q_{\phi}(z,y|t,x,c)||p_{\theta}(z,y|c)\big]\\
    &= \mathbf{E}_{q_\phi(y|t,x,c)} \operatorname{KL} \big[q_{\phi}(z|t,x,c,y)||p_{\theta}(z|c,y) \big] \\
    &+ \operatorname{KL} \big[q_{\phi}(y|t,x,c)||p_{\theta}(y|c) \big].
\end{split}
\end{equation}
Using this in equation \ref{eq:l_c} gives us the variational lower bound for causal CVAE as
\begin{equation}
\begin{split}
    &\mathcal{L}(\theta, \phi;t,x,c,y) =  \mathbf{E}_{q_\phi(z,y|t,x,c)} \log p_{\theta}(t|x,c,z,y) \\ 
    &+ \mathbf{E}_{q_\phi(z,y|t,x,c)} \log p_{\theta}(x|c,z,y) \\  
    &- \mathbf{E}_{q_\phi(y|t,x,c)} \operatorname{KL} \big[q_{\phi}(z|t,x,c,y)||p_{\theta}(z|c,y) \big] \\
    &- \operatorname{KL} \big[q_{\phi}(y|t,x,c)||p_{\theta}(y|c) \big]
\end{split}
\end{equation}

\begin{table*}[t]
\small
    \centering
    \begin{tabular}{l|cc|cc}
        Feature $\downarrow$ & \multicolumn{4}{c}{Average Treatment Effect} \\
        \hline
        Dataset $\rightarrow$ & \multicolumn{2}{c|}{Webhose} & \multicolumn{2}{c}{NYT} \\
        \hline
        Metric $\rightarrow$ & Participation & Replies & Comment & Upvote  \\
        \hline
        Word count & $-0.3816$ & $-0.1034$ & $-0.1034$ & $-0.0171$\\
        Paragraph count & $0.0079$ & $0.0038$ & $0.0025$ & $0.0078$\\
        Sentence count & $1.2308$ & $1.4453$ & $0.0203$ & $-0.0498$\\
        Images Count & NA & NA & $0.0279$ & $0.0387$ \\
        Links Count & NA & NA &  $-0.0459$ & $-0.0225$\\
        Slideshow Count & NA & NA & $0.0456$ & $-0.0077$\\
        Noun count & $-1.4758$ & $-0.1589$ & $-0.0062$ & $-0.0239$\\
        Verb count & $0.1591$ & $-0.8179$ & $0.0386$ & $0.0214$\\
        Adjective count & $-0.2364$ & $0.9527$ & $-0.0012$ & $-0.0008$\\
        Adverb count & $-0.0372$ & $-0.0372$ & $-0.0173$ & $-0.0037$\\
        Pronoun count & $-0.01949$ & $0.0203$ & $-0.0069$ & $-0.0153$\\
        \hline 
    \end{tabular}
    \caption{Average Treatment Effect of various article features on Comment count and Upvotes count for Webhose and NYT data}
    \label{tab:ate}
\end{table*}
\section{Conditional generation in Transformer}
\label{apx:transformer}

As discussed in section 4.3, we modify attention and normalization layers in a transformer architecture for adding metric as a guide. Inspired by \citet{zeng2020style}, we introduce the metric as follows:\\
\textbf{(1) Input embedding:} The metric control $y$ is directly added to the token and position embeddings of the input to the first transformer layer. This enables control by slanting the input representation towards the target metric.\\
\textbf{(2) Self-attention:} In self-attention mechanism of transformers, each input token is weighted with respect to other positions in the input. For each token $x_t$, query $q_t$, key $k_t$ and value $v_t$ is calculated using learned weight matrices $W^Q$,  $W^K$ and  $W^V$ respectively. The attention score for token $x_t$ is computed by a compatibility function of the corresponding query $q_t$ with the keys $k_i$ of other tokens and the attention vector is computed as the weighted average of these attention scores with the value vector $v_t$. This can be written as
\begin{equation}
    \operatorname{softmax}\bigg(\frac{QK^T}{\sqrt{d_k}}\bigg)V,
\end{equation}
where $d_k$ is the dimension of the key vector $k_t$. We modify this attention calculation to introduce the control $y$ by changing the query vector in the above equation to $q_t=\eta_t(y)$, where $\eta_t$ denoted an affine transformation. Modifying the query vector according to the specific target metric allows for biasing attention weights towards the target and capturing target control in the context representation, which aids in targeted decoding and generation.\\
\textbf{(3) Layer Normalization:} Classically, the layer normalization in transformers is calculated as
\begin{equation}
    \operatorname{LayerNorm}(\nu)=\gamma \frac{\nu-\mu}{\sigma}+\beta,
\end{equation}
where $\mu$ and $\sigma$ are the mean and standard deviation of the elements in $\nu$ and $\gamma$ and $\beta$ are the scale and bias parameters. The metric control, $y$, is used to modulate hidden representations of the generative model via normalization layers. The scale and bias parameters in the layer normalization are replaced as functions of $y$, namely $\gamma(y)$ and $\beta(y)$ in the above equation. As discussed in \citet{park2019normalisation}, normalization layer applied on input with same target control would \textit{wash away} the target information captured in the input to normalization layer. Adding target control in the scale and bias parameter ensures that the control is preserved through the normalization layers of transformer.

\noindent\textbf{Training details. } For fine-tuning, we prepend the input sentence with \textit{metric} identifiers, to keep the input layer unchanged. We, then, extract the prepended metric token and use it to modify attention and normalization layers as described earlier. The output of final transformer layer is fed into a pre-trained fastText model to estimate the fitment of generated text to the target metric class in the form of metric loss.\footnote{The computing infrastructure and hyper-parameter details are included in Appendix E} During inference, the generation is conditioned on the prompt, which is a combination of the topic and keywords. During training, the keywords and topic for the article is prepended to the input along with a \textit{\{start of text\}} token. Thus, the input is \textit{\{metric token\}+\{topic\}+\{start of keyword token\}+\{keywords\}+\{start of text token\}+\{article text\}}. The keywords and topics are available for the NYT dataset for each article, and are extracted from input text using topic modeling \cite{blei-2003-lda} as described in next section.

\section{Data Processing}
\label{apx:data}
\textbf{Webhose Covid-19 Dataset:} We use the Webhose dataset available at https://webhose.io/free-datasets/news-articles-that-mention-corona-virus/ that has $410,120$ data points in total. We choose the subset of this dataset limited to English. To remove any outliers, we heuristically choose articles with word count more than $30$ but less than $5000$ words in the article. The data contains engagement on various news articles in form of participation count, replies count and various other social media likes and share metrics. The social media metrics includes PinInterest, LinkedIn, Google+ shares and like, shares and comments on Facebook. Most of these are very sparse in the dataset, for instance, less than $\sim 12k$ data points have Facebook comments as non-zero. Thus, we choose participation count and replies count as good indicators to the engagement on the article and use these as our target metrics. We consider only the articles with participation count $>1$, leaving us with $39192$ data points in total. The metric value for participation count and replies count vary from $1-297$ and $0-5751$ respectively with a mean and standard deviation of $14.37, 27.90$ and $129.91, 446.71$. To control for these metrics in our models, we convert these to categorical variable with the threshold of $2$ and $21$ for participation count. The low bucket is the largest bucket with least standard deviation in the value of metric; the medium and high categories have almost same number of data points as shown in Table 1 in the paper. Similarly for replies count, the threshold is $2$ and $32$ with equal size of medium and high categories.

As mentioned earlier, the context for generative models includes keywords and topic of the article, that acts as ``prompt" during inference stage. For webhose data, the keywords are not directly available in the dataset, NYT-comments dataset has keywords. We extract the keywords as top $n$ ($n=10$) words from the articles using TF-IDF vectors. The topics are extracted by topic modeling using Latent Dirichlet Allocation (LDA) \cite{blei-2003-lda}. We choose $20$ topics with a seed of $23$ and then represent the topic of each input article as the corresponding topic identifier ranging from $1$-$20$. For transformer-based model, the keyword and topic tokens are added to the pre-trained tokenizer.

\section{Causal Features}
The various textual features considered for causal effect are as listed in Table \ref{tab:ate}. The average treatment effect on NYT data metrics -- Comment count and Upvote count is as shown. Here, the significance level is empirically chosen as $0.01$. Thus, features with $|$ATE$|>0.01$ on comment count or upvote count $y$ are included in the corresponding causal generative model. For Webhose data, we choose significance level of $0.1$ and consider features with ATE of greater than $0.1$ in magnitude as `causally significant' features.

\section{Reproducibility checklist}
\subsection{Hyper-parameters}
The causal feature identification models are trained on a train-test split of 90-10, using a random seed $23$ with stratified sampling over the outcome values, for over $10$ epochs in batches of size of $5$.

For transformers, we use HuggingFace\footnote{https://github.com/huggingface/transformers} implementation of GPT-2 and make the model and training changes as described in the paper. The hyper-parameters are kept the same as the original implementation for uniformity. For the loss term mentioned in equation 11 of the paper, we set $\lambda_{G}$, $\lambda_{metric}$, $\lambda_{causal}$ as $1$. We train these models with a batch size of $2$ for over $3$ epochs. The training time over $4$ GPUs was about $14$ hours for webhose data and about $5$ hours for NYT dataset.

For the CVAE model, we use adam optimizer. We initiate the training with the learning rate of $0.001$ with learning rate decay of $0.6$. We train the models over $30$ epochs with an early stopping criterion of $0.996$ threshold. 

\subsection{Resources}
All the training experiments were run on a 4 GPU machine with 64-bit 16 core tesla v100 processor and 100 GB RAM.